\pdfoutput=1

\documentclass[11pt]{article}

\usepackage{emnlp2021}

\usepackage{times}
\usepackage{latexsym}

\usepackage[T1]{fontenc}

\usepackage[utf8]{inputenc}

\usepackage{microtype}

\usepackage{url}
\usepackage{graphicx}
\usepackage{wrapfig}
\usepackage{amsmath}
\usepackage{amssymb}
\usepackage{color,xcolor,colortbl}
\usepackage{enumitem}
\usepackage{algorithm}
\usepackage{algorithmic}
\usepackage{bm,bbm}
\usepackage{booktabs}
\usepackage{mathtools}
\usepackage{array}
\usepackage{multirow}
\usepackage{soul}
\usepackage{subcaption}
\usepackage{pbox}
\usepackage{pifont}
\usepackage{arydshln}
\usepackage{boldline}
\usepackage{dsfont}
\usepackage{comment}
\usepackage[scaled=0.95]{beramono}

\DeclareMathOperator*{\argmax}{arg\,max\,}
\newcommand\numberthis{\addtocounter{equation}{1}\tag{\theequation}}

\definecolor{darkblue}{rgb}{0.0, 0.0, 0.55}
\setul{0.5ex}{0.3ex} 
\setulcolor{blue} 
\usepackage{pifont}
\newcommand{\done}{\rlap{$\square$}{\raisebox{-0.5pt}{\large\hspace{-3pt} \textsf{x}}}%
\hspace{2pt}}

\definecolor{darkblue}{rgb}{0.0, 0.0, 0.55}
\definecolor{midnightblue}{HTML}{191970}
\definecolor{darkgreen}{HTML}{006400}
\definecolor{red}{HTML}{ff0000}
\definecolor{gold}{HTML}{ffd700}
\definecolor{mediumvioletred}{HTML}{c71585}
\definecolor{lime}{HTML}{00ff00}
\definecolor{aqua}{HTML}{00ffff}
\definecolor{fuchsia}{HTML}{ff00ff}
\definecolor{lightpink}{HTML}{ffb6c1}
\definecolor{dodgerblue}{HTML}{1e90ff}
\definecolor{deepskyblue}{HTML}{00bfff}
\definecolor{deeppink}{HTML}{ff1493}
\definecolor{orangered}{HTML}{ff4500}
\definecolor{mediumseagreen}{HTML}{3cb371}
\definecolor{saddlebrown}{HTML}{8b4513}

\title{StreamHover: Livestream Transcript Summarization and Annotation}

\author{Sangwoo Cho,$^\dagger$ Franck Dernoncourt,$^\ddagger$ Tim Ganter,$^\ddagger$ Trung Bui,$^\ddagger$ Nedim Lipka,$^\ddagger$\\
\textbf{Walter Chang,$^\ddagger$ Hailin Jin,$^\ddagger$ Jonathan Brandt,$^\ddagger$ Hassan Foroosh,$^\dagger$ Fei Liu$^\dagger$}
\\[0.5em]
$^\dagger$University of Central Florida \quad
$^\ddagger$Adobe Research
\\[0.3em]
\texttt{\footnotesize swcho@knights.ucf.edu \quad \{foroosh,feiliu\}@cs.ucf.edu}\\
\texttt{\footnotesize \{dernonco,timganter,bui,lipka,wachang,hljin,jbrandt\}@adobe.com}\\
}

\date{}

\begin{document}
\maketitle
\begin{abstract}

With the explosive growth of livestream broadcasting, there is an urgent need for new summarization technology that enables us to create a preview of streamed content and tap into this wealth of knowledge. 
However, the problem is nontrivial due to the informal nature of spoken language. 
Further, there has been a shortage of annotated datasets that are necessary for transcript summarization.
In this paper, we present StreamHover, a framework for annotating and summarizing livestream transcripts.
With a total of over 500 hours of videos annotated with both extractive and abstractive summaries, our benchmark dataset is significantly larger than currently existing annotated corpora.
We explore a neural extractive summarization model that leverages vector-quantized variational autoencoder to learn latent vector representations of spoken utterances and identify salient utterances from the transcripts to form summaries.
We show that our model generalizes better and improves performance over strong baselines.
The results of this study provide an avenue for future research to improve summarization solutions for efficient browsing of livestreams.

\end{abstract}



\section{Introduction}

One of the most powerful communication mediums is livestreaming.
New platforms such as YouTube Live, Twitch, Instagram Live and TikTok encompass a variety of topics, ranging from video games to social media to professional sports.
We are particularly interested in livestreams that are distinguished by three characteristics:
\emph{Excessive length}, the recordings could last from several minutes to several hours;
\emph{Verbal communication}, the use of natural language is the primary means of communication, in contrast to gestures or facial expressions;
\emph{Informal nature}, the streamers' language is mostly informal and unplanned, unlike news broadcasts.
Without an effective mechanism to summarize such streamed content, livestreaming platforms may not fully meet the needs of their customers.

\begin{figure}
\centering
\includegraphics[width=3.1in]{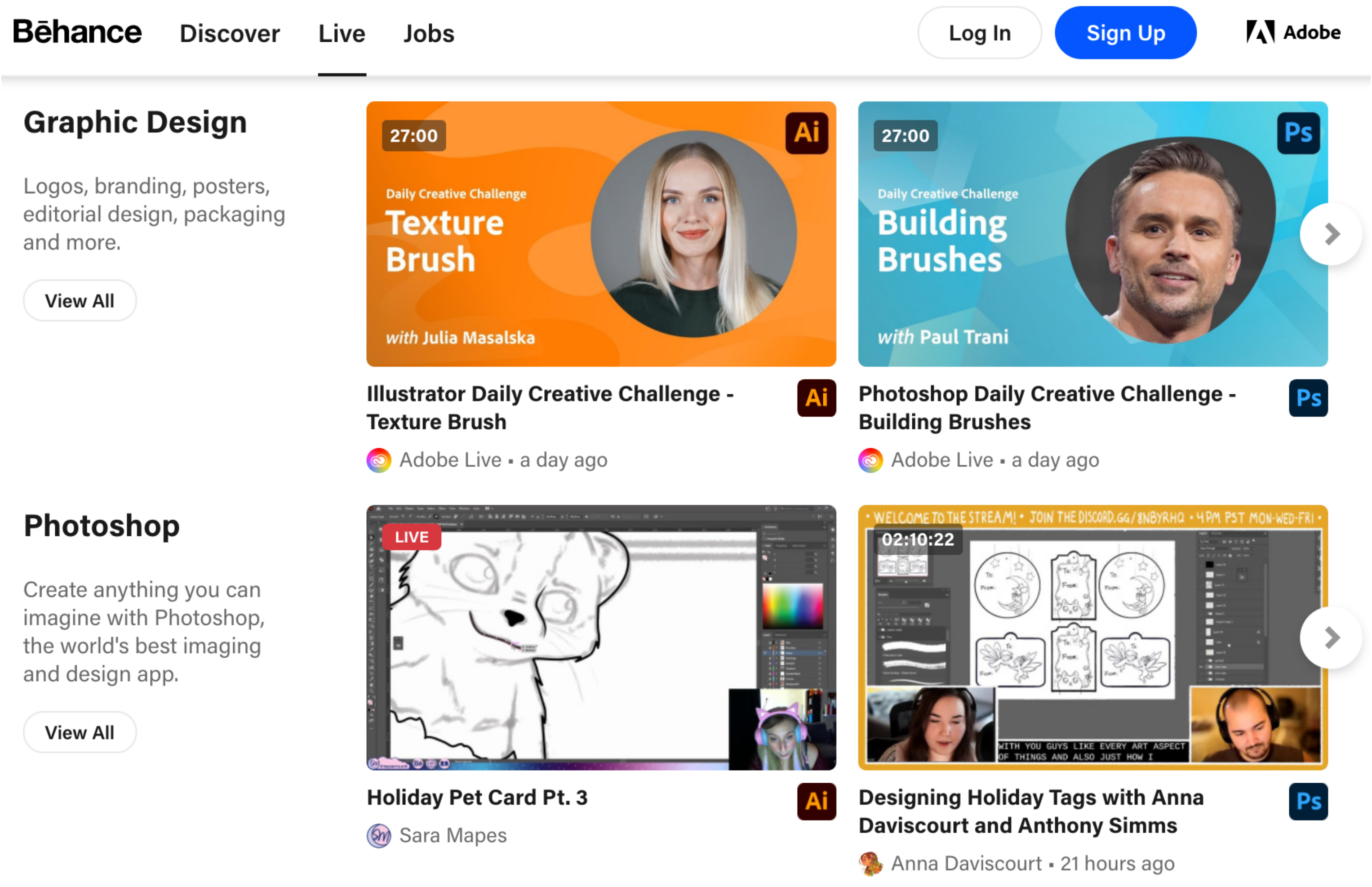}
\caption{An example of streamed content on B\={e}hance, a streaming platform for artists and designers to showcase creative work related to Adobe Photoshop, Illustrator, Fresco, UI/UX, photography and more.
\textsc{Top}: The videos are each 27 minutes long.
\textsc{Bottom}: One video is being broadcast live, the other is $>$2 hours long.
}
\label{fig:behance}
\vspace{-0.15in}
\end{figure}

Our goal in this work is to create a \emph{text preview} of the streamed content. 
When a user \textbf{hovers over} the thumbnail or scrolls past a video, they are shown a preview of the content.
We present a dataset of over 500 hours of video footage, which were streamed live on a social media platform (\textsf{\small behance.net}) created to showcase and discover creative work.
Figure~\ref{fig:behance} shows an example of the streams, where the artists showcase the use of Adobe Photoshop and Illustrator in designing holiday cards and posters. 
It is necessary to point out that video analysis is not suitable here, as the video only mirrors the artists' screen content.
As a first step towards automatic creation of a text preview, we focus on identifying salient utterances to produce an extract from the livestream transcript.

We make use of vector-quantized variational autoencoders (VQ-VAE; van den Oord et al., 2017)\nocite{NIPS2017_7a98af17} to identify salient utterances.
The model has been applied successfully to opinion summarization that learns in-domain sentence representations~\cite{angelidis2020extractive}, which is essential for adaptation of general-domain models.
We refrain from using sequential methods for utterance selection. 
First, it is difficult to scale up sequential prediction to process transcripts that exceed the maximum allowed length, even with models that handle long text~\cite{beltagy2020longformer,Zhao2020Transformer-XH:}.
Second, sequential methods~\cite{narayan-etal-2018-ranking,xiao-carenini-2019-extractive} may not give enough flexibility to select salient utterances \emph{on-the-fly} when content is being streamed live, thus they are unsuitable for our case.


There has been a shortage of annotated datasets that are necessary for livestream transcript summarization.
We build a browser-based user interface for summary annotation that provides to the annotators a clip of the livestream recording alongside a synchronized display of the transcript.
The interface allows annotators to conveniently label summary utterances and write an abstractive summary using their own words (Figure~\ref{fig:interface}).
With a total of 500 hours of annotated video footage, our dataset is notably larger than existing annotated corpora for transcript summarization~\cite{1198793,10.1007/11677482_3}. 
We compare our summarization approach with strong baselines on the dataset and shed light on the task of livestream transcript summarization. 
Our contributions are as follows.
\begin{itemize}[topsep=5pt,itemsep=0pt,leftmargin=*]

\item We create a detailed annotation interface and new benchmark dataset for automatic summarization of livestream transcripts.
An informative preview of streamed content is of crucial importance to users when considering whether to hit play.

\item We present \textbf{StreamHover}, a unsupervised model based on VQ-VAE to identify salient utterances from livestream transcripts to form preview summaries. 
We evaluate the method across multiple dimensions and discuss its strengths and weaknesses.
Empirical results show that our method outperforms strong summarization baselines.\footnote{Our annotations and source code are available at \url{https://github.com/ucfnlp/streamhover}}


\end{itemize}



\begin{figure}
\centering
\includegraphics[width=3in]{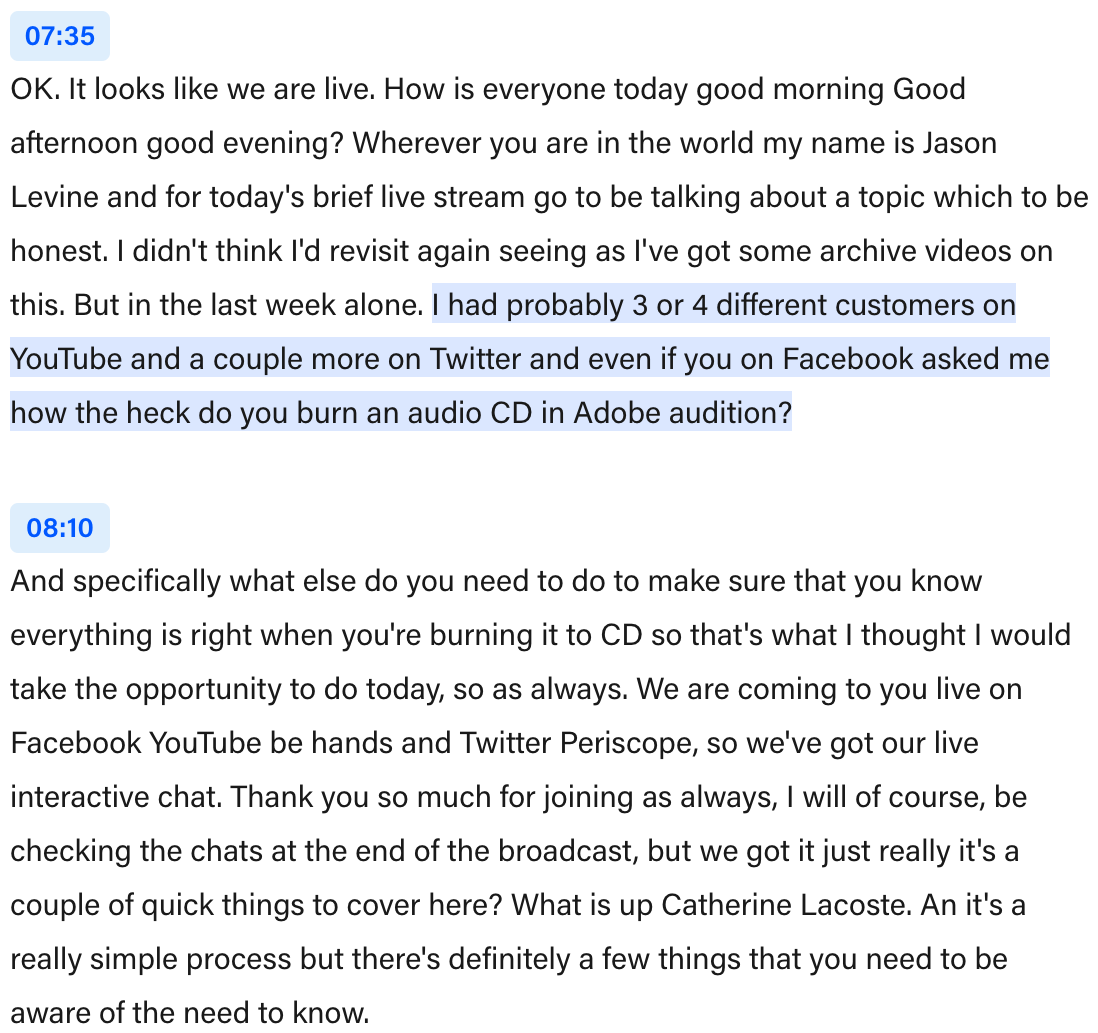}
\caption{A transcript snippet from ``\emph{How to Create an Audio CD in Adobe Audition}.''
Most utterances are off-topic, except for the one marked in blue, suggesting the information density of livestream transcripts is low. 
}
\label{fig:snippet}
\vspace{-0.1in}
\end{figure}

\section{Related Work}
\label{sec:related}

Closed captions are often provided onscreen, turning streaming videos into text on an unprecedented scale~\cite{google2020}.
However, there are very few summarization studies that attempt to generate text previews of streaming videos to help users browse or refind information that has been watched before. 
Neural text summarizers have focused primarily on written text, including news articles, reviews, scientific papers and book chapters~\cite{see-etal-2017-get,tan-etal-2017-abstractive,chen-bansal-2018-fast,narayan-etal-2018-dont,gehrmann-etal-2018-bottom,cohan-etal-2018-discourse,liu-lapata-2019-text,fabbri-etal-2019-multi,brazinskas-etal-2020-shot,ladhak-etal-2020-exploring,song-etal-2021-new}.
Despite their success, it remains unclear as to if and how the summarizers can be extended to spoken text, whose utterances may have very low information density.



It is crucial to identify salient content from transcripts where a substantial number of utterances are devoted to informal chit-chats in an attempt to connect with the audience (Figure~\ref{fig:snippet}).
We investigate extractive rather than abstractive approaches as the latter are prone to generate hallucinated content that does not exist in the source text~\cite{cao-etal-2017-fsum,kryscinski-etal-2019-neural,lebanoff-etal-2019-analyzing,maynez-etal-2020-faithfulness}.
The problem could be exacerbated by ungrammatical spoken utterances and transcription errors.
Instead, we consider VQ-VAE, an unsupervised representation learning technique~\cite{NIPS2017_7a98af17,jin-etal-2020-discrete,angelidis2020extractive} for content extraction. 
Unsupervised training of the VQ-VAE model and its inference could potentially be performed at the same time,
allowing important utterances to be extracted from a transcript segment \emph{on-the-fly} during streaming, without interrupting the learning process. 
It is also easier to tailor the model to specific domains compared to contemporary extractive methods~\cite{yasunaga-etal-2017-graph,dong-etal-2018-banditsum,xu-durrett-2019-neural,wang-etal-2020-heterogeneous}.

Our work contributes to a refined understanding of transcript summarization, which is understudied relative to its importance and potential. 
The transcripts may be obtained from channels such as movies and TVs~\cite{papalampidi-etal-2020-screenplay,DBLP:journals/corr/abs-2104-07091}, interviews~\cite{zhu-etal-2021-mediasum}, multi-party meetings~\cite{murray-carenini-2008-summarizing,wang-cardie-2013-domain2,li-etal-2019-keep,koay-etal-2020-domain,koay-etal-2021-sliding,zhong2021}, telephone speech~\cite{kafle-huenerfauth-2018-corpus} and more.
The main thrust distinguishing our work with others is the combination of a benchmark summarization dataset, novel summarization methods and a challenging new domain where salient content is scattered throughout the transcript and mixed with substantial chit-chats.
We do not make use of video event detection or multi-modal fusion~\cite{zhu-etal-2018-msmo,palaskar-etal-2019-multimodal,li-etal-2020-vmsmo} as little information could be gleaned from videos that mirror the artists' desktop.
Instead, we focus on generating short descriptions from transcripts and leave for future work cross-modality research.
We describe our data annotation process in the following section.

\section{Our Dataset}
\label{sec:data}


We aim to create a large and representative corpus containing transcripts and summaries of streamed videos. 
We explore a leading social media platform (\textsf{\small Behance.net}) supported by Adobe Creative Cloud that features livestreams of creative work by artists and designers. 
The website boasts over 10 million users, who watch artists and designers livestream when they create. 
Our data are extracted from this website, containing a large quantity of streamed videos (>5,000), the length of which ranges from minutes to several hours.
The streamers' language is unplanned, instead of rehearsed as that of TED talks~\cite{Hernandez_2018}.

\begin{figure}[t]
\centering
\includegraphics[width=3.1in]{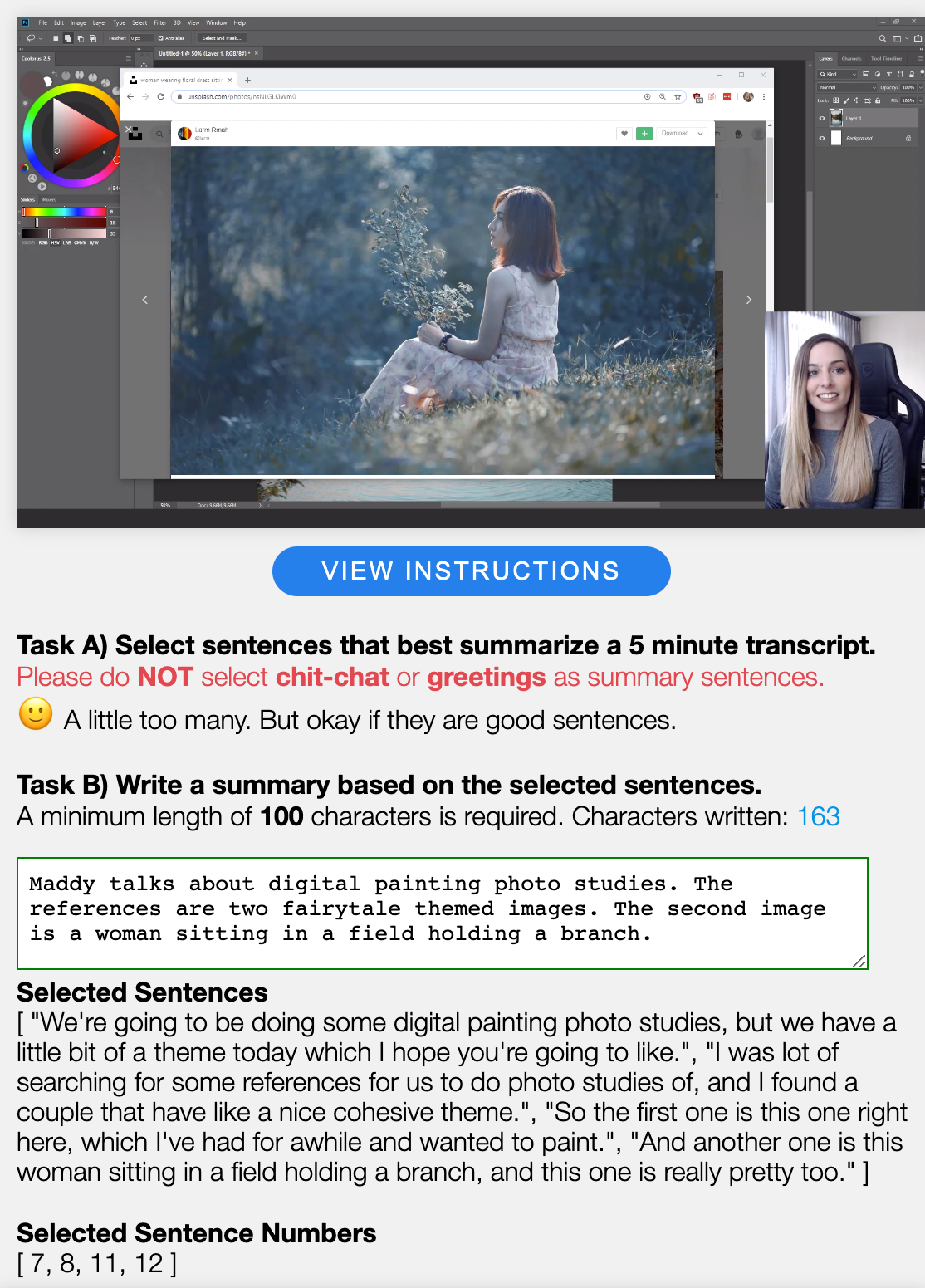}
\caption{
An example of our browser-based annotation interface. 
It includes a clip of the streamed video alongside a display of the transcript (omitted for space).
The streamer talks about \emph{Digital Painting with Maddy Bellwoar} to create fairytale themed images.
The annotators are asked to write a concise summary of this clip using their own words (Task A) and identify summary utterances (Task B).
}
\label{fig:interface}
\vspace{-0.1in}
\end{figure}



We obtain a total of 5,398 streamed videos.
The metadata of a video includes its ID, duration, title, a short description and the transcript.
Automatic transcription was provided by Microsoft Automatic Speech Recognition which helps make videos accessible to a wide audience.
Each transcript contains a set of segments, each corresponds to about 30 seconds of audio. 
Each segment contains a set of utterances.\footnote{An utterance is the smallest unit of speech in improvised spoken language.
It usually begins and ends with a clear pause. 
We obtain utterance boundaries from Microsoft's ASR system.} 
Figure~\ref{fig:snippet} shows an example of the segments and utterances.
The offset of the segment indicates the number of minutes since the beginning of the recording.

When a user hovers over the thumbnail or scrolls past a video, we expect a textual summary to give a glimpse of the verbal content.
This view of summarization leads us to annotate salient content across the video in an \emph{equally detailed} manner.
It naturally avoids lead bias that is ubiquitous in news~\cite{grenander-etal-2019-countering}.
We segment a video into 5-minute clips and annotate each clip for summary-worthy content.
A clip contains an average of 51 utterances and 460 words.
Due to time and budget constraints, we select 370 streamed video for summary annotation.\footnote{
Details of video selection are provided in Supplementary.
}
Table~\ref{tab:datasets} provides a detailed comparison of our annotated corpus with previous datasets,
including Switchboard~\cite{225858}, ICSI~\cite{1198793} and AMI~\cite{10.1007/11677482_3} that contain both transcripts and human-annotated extractive/abstractive summaries.
With a combined duration of 500 hours, our dataset is substantially larger than previously released datasets.

\begin{table}[t]
\setlength{\tabcolsep}{5pt}
\renewcommand{\arraystretch}{1.1}
\centering
\begin{footnotesize}
\begin{tabular}{|l|l|l|c|c|}
\hline
\textbf{Dataset} & \textbf{Type} & \textbf{Duration} & \textbf{Extr.} & \textbf{Abst.} \\
\hline
\hline
Switchboard & Telephone & 300 Hrs & No$^\dagger$ & No \\
ICSI  & Meeting & 75 Hrs & Yes & Yes \\
AMI  & Meeting & 100 Hrs & Yes & Yes \\
\textbf{StreamHover} & \textbf{Livestream} & \textbf{500 Hrs} & \textbf{Yes} & \textbf{Yes}\\
\hline
\end{tabular}
\end{footnotesize}
\vspace{-0.05in}
\caption{A comparison of the transcript summarization datasets
with manually annotated extractive/abstractive summaries.
``Yes/No'' indicate a summary type is available or not.
$\dagger$ suggests only small pilot summary annotations are available for the Switchboard dataset~\cite{penn-zhu-2008-critical}.
With a total duration of over 500 hours, our dataset is notably larger than similar datasets.
}
\label{tab:datasets}
\end{table}



We recruit 12 workers from \textsf{\small Upwork.com} and validate their language skills for summary annotation. 
Upwork is a freelancing platform that allows us to reach out to workers directly to ensure our instructions are fully understood.
Each worker is asked to write a concise summary for a given clip using their own words (Task A) and identify summary utterances (Task B) using the graphical user interface (Figure~\ref{fig:interface}), which shows a clip of the streamed video alongside a synchronized display of the transcript.
Additionally, our guidelines suggest a good summary of Task A should have at least 100 characters and that of Task B should have between 50 and 80 words ($\sim$15\% compression).
As is the case with meeting summaries~\cite{1198793}, a clip is annotated by a single worker owing to an expensive annotation process. 
The worker can also identify a clip to be chitchat, in which case it will not be annotated for summaries.

Table~\ref{tab:data} shows our dataset statistics.
On average, a human \emph{abstract} contains 3 sentences (36 words) and a human annotated \emph{extract} contains 5.5 utterances (80 words).
Moreover, summary utterances constitute 8.9\% and 8.6\% of all utterances in terms of number and duration.
We study inter-annotator agreement by inviting three workers to each annotate 8 hours of video that contains a total of 96 clips. 
Using 10-second intervals as measuring units,\footnote{
We use 10-second intervals rather than utterances as measuring units as the duration of utterances vary.
If annotators all selected some content, or no content at all, from a 10-second interval, they are in agreement.
} 
the Fleiss' Kappa on identifying summary utterances is 0.13.
We note that the score is within the range of what is normally found in annotating speech transcripts for extractive summaries (0.1$\sim$0.3; Marge et al., 2010)\nocite{marge-etal-2010-using}, as annotating spoken text is a highly challenging task.
We find that annotators tend to perceive the same region as salient but they may disagree as to which utterances should be included in the summary due to verbosity of spoken text.
We refer the reader to~\cite{artstein2008} for interpretations and improvements to IAA.




\begin{figure*}
\centering
\includegraphics[width=5.3in]{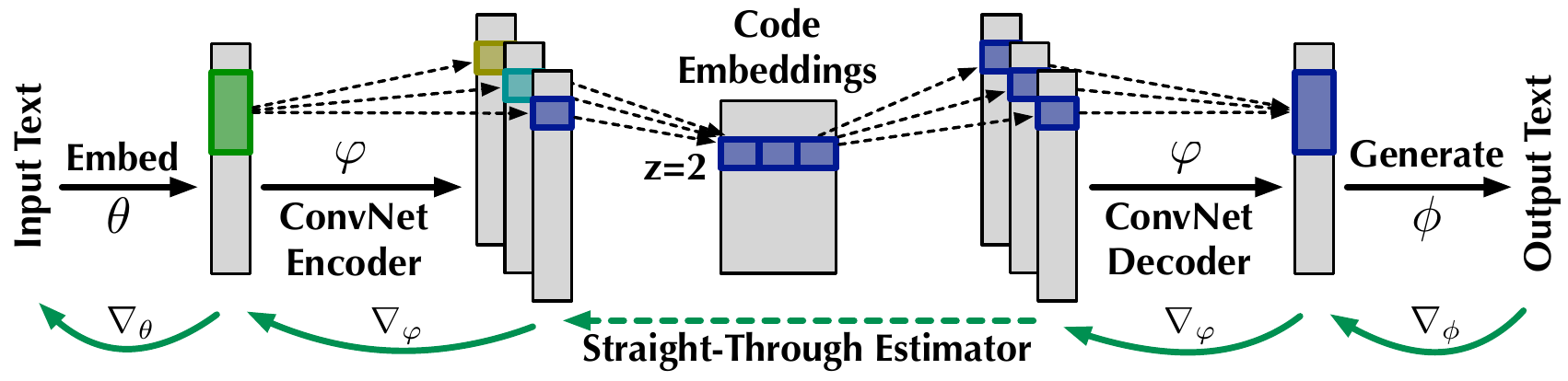}
\caption{Our summarizer embeds an input utterance using BERT, transforms BERT's semantic space to a set of latent codes, then reconstructs the utterance using the code embeddings. We identify summary utterances as those associated with prominent latent codes/topics. The model is trained using a dictionary learning algorithm for code embeddings ($\mathbf{E}$) and backpropagation with a straight-through estimator for model parameters $\theta$, $\varphi$ and $\phi$.
}
\label{fig:architecture}
\end{figure*}

\begin{table}[t]
\setlength{\tabcolsep}{4pt}
\renewcommand{\arraystretch}{1.15}
\centering
\begin{small}
\begin{tabular}{|lr|}
\hline
\multicolumn{2}{|l|}{\textbf{The Bēhance Corpus}}\\ 
\hline
\hline
Total number of annotated videos & 370\\
Total annotated duration in hours & 500\\ 
Total number of utterances & 331,928\\
Average number of utterances in a clip & 61.23\\
Average duration of utterances in seconds & 3.04\\
Average number of words in an utterance & 9.48\\
Total number of annotated clips & 6,003\\
Total number of chitchat clips & 582\\
Total number of human annotators & 12\\
Avg. \# of (sentences) words in a human abstract & (3.0) 36\\
Avg. \# of (utterances) words in a human extract & (5.5) 80\\
Percentage of duration of summary utterances & 8.57\%\\
\hline
\end{tabular}
\end{small}
\vspace{-0.05in}
\caption{Statistics of our dataset.
}
\label{tab:data}
\vspace{-0.05in}
\end{table}

\section{Summarization}
\label{sec:approach}

Let $\mathcal{X}$ denote a sequence of spoken utterances from a segment of the transcript.
Our summarizer aims to extract a subset of utterances $\mathcal{Y} \subset \mathcal{X}$ that convey the essential content of the input.
We experiment with an unsupervised summarizer that leverages vector-quantized variational autoencoders (VQ-VAE; van den Oord et al., 2017)\nocite{NIPS2017_7a98af17} to learn utterance representations and identifies summary utterances.
The method was explored for opinion summarization~\cite{angelidis2020extractive} and machine translation~\cite{prato-etal-2020-fully}.
We are interested in using the method to account for domain characteristics of livestreams, which showcase new and creative work of artists and designers on their use of Photoshop, Illustrator, and other tools.\footnote{A show may have more than one host, their utterances are treated indistinguishably due to speaker diarization that identifies different speakers in the audio is not provided.}

VQ-VAE is a powerful framework for learning latent variable models using deep neural networks. 
It learns \emph{discrete} vector representations for an utterance, which is then used to categorize the utterance along various dimensions. E.g., ``\emph{Good morning Hi Everybody}'' suggests a greeting and opens up a dialogue; ``\emph{I had probably 3 or 4 different customers on YouTube and ... on Facebook asked me how the heck do you burn an audio CD in Adobe Audition}'' engages the audience and introduces the main topic.
The VQ-VAE method groups utterances based on their discrete representations and selects salient utterances to form a summary.

We employ an embedding function $\texttt{\small Embed}_\theta(\cdot)$ to map an input utterance $x$ into a semantically meaningful space.
The space is subsequently discretized according to a codebook.
To achieve this, we prefix $x$ with a \texttt{\small [CLS]} token and append a \texttt{\small [SEP]} token, pass it into a BERT model, then obtain the vector corresponding to \texttt{\small [CLS]} as a pooled representation of the utterance, denoted by $\mathbf{h} \in \mathbb{R}^H$ (Eq.~(\ref{eq:embed})).
We use a $\texttt{\small ConvEncoder}_\varphi(\cdot)$ with a set of $D$ filters to convolve the input $\mathbf{h}$.
The output is a sequence of feature vectors $[\mathbf{q}_1, \cdots, \mathbf{q}_H]$ where $\mathbf{q}_i \in \mathbb{R}^{D}$ (Eq.~(\ref{eq:conv_encoder})).
We define a codebook $\mathbf{E} = [\mathbf{e}_1, \cdots, \mathbf{e}_K]$, where $K$ is the number of latent codes and $\mathbf{e}_k \in \mathbb{R}^D$ is the $k$-th code embedding.
The $i$-th feature $\mathbf{q}_i$ is assigned to the latent code $z_i$ whose embedding $\mathbf{e}_{z_i}$ has the minimum Euclidean distance with it (Eq.~(\ref{eq:z_i})).
Our method essentially discretizes the $H$-dimensional semantic space by producing latent codes $\{z_i\}_{i=1}^H$, one for each dimension of the semantic space. 
\begin{align*}
&\mathbf{h} = \texttt{\small Embed}_\theta(x) \in \mathbb{R}^H
\numberthis\label{eq:embed}\\
&[\mathbf{q}_1, \cdots, \mathbf{q}_H] = \texttt{\small ConvEncoder}_\varphi(\mathbf{h}), \mathbf{q}_i \in \mathbb{R}^{D}
\numberthis\label{eq:conv_encoder}\\
&z_i = \argmax_k -\lVert\mathbf{q}_i - \mathbf{e}_k\rVert_2, \,\, i \in [H] 
\numberthis\label{eq:z_i}
\end{align*}

With the latent code embeddings $[\mathbf{e}_{z_1}, \cdots, \mathbf{e}_{z_H}]$, we seek to reconstruct the input utterance, 
which is achieved by generating a dense vector $\mathbf{\widetilde{h}}$ using a $\texttt{\small ConvDecoder}_{\varphi}(\cdot)$ (Eq.~(\ref{eq:conv_decoder})).
$\mathbf{\widetilde{h}}$ is then fed to a Transformer decoder to reconstruct the original utterance $\widetilde{x}$ (Eq.~(\ref{eq:generate})). 
In this process, the code embeddings serve as ``topic vectors'' that group dimensions of the semantic space into clusters relevant to the application domain.
Our model parameters include those used by the BERT encoder and Transformer decoder ($\theta$ and $\phi$),
the convolutional encoder and decoder that use tied parameters ($\varphi$),
and embeddings of the codebook $\mathbf{E}$. 
\begin{align*}
\mathbf{\widetilde{h}} &= \texttt{\small ConvDecoder}_{\varphi}([\mathbf{e}_{z_1}, \cdots, \mathbf{e}_{z_H}]) \in \mathbb{R}^H
\numberthis\label{eq:conv_decoder}\\
\widetilde{x} &= \texttt{\small Generate}_\phi(\mathbf{\widetilde{h}})
\numberthis\label{eq:generate}
\end{align*}

We next describe the loss function used to learn these parameters. 
The loss function of our model comprises of three parts, including
a cross-entropy term between the original and reconstructed utterance $\texttt{\small XEnt}(x, \widetilde{x})$ that optimizes the BERT embedder $\theta$, Transformer generator $\phi$, and convolutional encoder and decoder $\varphi$, as shown in Figure~\ref{fig:architecture}.
The gradients will, however, bypass the latent code embeddings due to the straight-through estimator~\cite{bengio2013estimating}. 
To learn code embeddings in an end-to-end manner, we use a dictionary learning algorithm~\cite{NIPS2017_7a98af17} that moves code embeddings $\mathbf{e}_{z_i}$ towards feature vectors $\mathbf{q}_i$ by minimizing the $l_2$-distance between the two vectors $\lVert\mathbf{e}_{z_i} - \texttt{\small sg}(\mathbf{q}_i)\rVert_2^2$,
where $\texttt{\small sg}(\cdot)$ is a stop-gradient operator that constrains its operand to be a non-updated constant during backpropagation,
i.e., it stops $\mathbf{q}_i$ from being updated.
As illustrated in Eq.~(\ref{eq:loss}), we additionally apply a commitment loss to encourage the feature vector $\mathbf{q}_i$ to commit to a code embedding.
$\lVert\texttt{\small sg}(\mathbf{e}_{z_i}) - \mathbf{q}_i\rVert_2^2$ prevents $\mathbf{q}_i$ from deviating too much from the code embedding $\mathbf{e}_{z_i}$.
This loss term is associated with a coefficient $\beta \in [0,1]$.
\begin{align*}
\mathcal{L}(\theta) & = \texttt{\small XEnt}(x, \widetilde{x}) + \textstyle\sum_i \lVert\mathbf{e}_{z_i} - \texttt{\small sg}(\mathbf{q}_i)\rVert_2^2\\
& + \beta \textstyle\sum_i \lVert\texttt{\small sg}(\mathbf{e}_{z_i}) - \mathbf{q}_i\rVert_2^2
\numberthis\label{eq:loss}
\end{align*}

At test time, we define summary utterances as those associated with prominent latent codes/topics.
Given a set of $N$ utterances, we obtain latent codes from the $n$-th utterance using Eq.~(\ref{eq:z_i}), denoted by $\textstyle\{z_i^{(n)}\}_{i=1}^H$. 
This gives a total of $N \times H$ codes from which we find prominent ones. 
They are denoted by $\mathcal{P}$ which contains a set of most frequently occurring codes.
A score $\mathcal{S}(x_n)$ is assigned to utterance $x_n$ that computes how often it is associated with those prominent codes $\mathcal{P}$.
In Eq.~(\ref{eq:score_x}), $\textstyle\sum_{i=1}^H \mathds{1}[z_i^{(n)}=k]$ indicates the number of times the $n$-th utterance is assigned to code $k$, where $k$ belongs to $\mathcal{P}$.
Finally, we extract $K$ highest-scoring utterances to form an extractive summary of the input.
\begin{align*}
\mathcal{S}(x_n) = \textstyle\sum_{k \in \mathcal{P}} \textstyle\sum_{i=1}^H \mathds{1}[z_i^{(n)}=k]
\numberthis\label{eq:score_x}
\end{align*}

Our method draws on the convolutional encoder and decoder to transform BERT's semantic space to map each dimension to a latent code.
The summary selection process is deterministic and our encoder takes full advantage of a large, pre-trained model to produce initial utterance representations. 
This design sets our method apart from that of Angelidis et al.~\shortcite{angelidis2020extractive}.
Moreover, the method has the potential for modelling topic transitions between utterances to improve summarization of livestreams, which we leave for future work.


\section{Experiments}
\label{sec:experiments}

\textbf{Dataset.}\quad 
Finding salient content from livestream transcripts is a ``needle-in-the-haystack'' problem.
Our summarization dataset contains a total of 370 videos split into short clips of 5 minutes each. 
The annotators manually annotated 5,421 clips ($\sim$451 hours) with extractive and abstractive summaries.
582 clips ($\sim$49 hours) are removed because they are identified to contain only chit-chats.
The dataset is divided into training, validation and test splits:
\begin{itemize}[topsep=4pt,itemsep=-1pt,leftmargin=*]
\item 3,884 clips (320 videos / 323 hours) in training,
\item 728 clips (25 videos / 61 hours) in validation,
\item 809 clips (25 videos / 67 hours) in test split.
\end{itemize}


We call our summarizer ``\textbf{StreamHover}.''
When a user hovers their mouse over a video's timeline, a summary preview is shown and keeps updating.
As a first attempt, StreamHover focuses on extracting salient utterances from individual clips instead of whole streams to encourage selected utterances to be mostly evenly distributed across the stream.
When the content is provided live, the stream can be divided into short clips and our algorithm consumes one clip at a time to produce summaries on-the-fly.
It is important to note that extracting summary utterances remains challenging even for modern neural summarizers.
E.g., Kedzie et al.~\shortcite{kedzie-etal-2018-content} reveal that summarizers may not effectively identify summary content without a dependency on intentional lead bias in news writing.
Our setting is challenging as not only are there few utterances deemed to be summary-worthy but such utterances can occur anywhere in a video clip.


\vspace{0.05in}
\noindent\textbf{Baselines.}\quad 
We compare StreamHover with state-of-the-art extractive and abstractive summarizers.
The abstractive summarizers generate an abstract from the transcript of a clip without tuning.\footnote{In a similar vein, our summarizer uses the transcripts to learn model parameters. It does not require utterance labels. 
}
These include {BART-large}, {BART-large-cnn}~\cite{lewis-etal-2020-bart} and {T5}~\cite{JMLR:v21:20-074},
which are some of the strongest performing neural abstractive summarizers that are pre-trained on language modeling and summarization tasks. 

The unsupervised extractive summarizers extract salient utterances from a clip.
{LexRank}~\cite{Erkan:2004} and {TextRank}~\cite{mihalcea-tarau-2004-textrank} are graph-based models that extract relevant sentences based on eigenvector centrality.
{SumBasic}~\cite{Vanderwende:2007} assigns higher scores to sentences containing frequently occurring content words.
We further compare to a novel unsupervised graph-based summarization method for speech transcripts:
{FluCovRank}~\cite{shang-etal-2018-unsupervised} groups utterances into clusters, generates an abstractive sentence from each cluster, then selects the best elements from abstractive sentences under a budget constraint.
Finally, we compare our approach with the Quantized Transformer~\cite{angelidis2020extractive}, which uses a clustering interpretation of the quantized space and two-step sampling algorithm to extract summary sentences from reviews.


\begin{table*}
\setlength{\tabcolsep}{4pt}
\renewcommand{\arraystretch}{1.15}
\centering
\begin{footnotesize}
\begin{tabular}{|l||rrr|r||rrr|r||rrr|r|}
\hline
& \multicolumn{4}{c||}{\textbf{3-Sentence Output}} & \multicolumn{4}{c||}{\textbf{4-Sentence Output}} & \multicolumn{4}{c|}{\textbf{5-Sentence Output}} \\
\textbf{System} & \textbf{P (\%)} & \textbf{R (\%)} & \textbf{F (\%)} & \textbf{\#Wrds} & \textbf{P (\%)} & \textbf{R (\%)} & \textbf{F (\%)} & \textbf{\#Wrds} & \textbf{P (\%)} & \textbf{R (\%)} & \textbf{F (\%)} & \textbf{\#Wrds}\\
\hline
\hline
LEAD-N & 18.83 & 9.57 & 12.5 & 38.53 & 18.63 & 12.61 & 14.77 & 51.35 & 18.71 & 15.76 & 16.82 & 64.04\\
SumBasic & 8.32 & 4.15 & 5.45 & 29.44 & 8.47 & 5.61 & 6.63 & 39.97 & 8.83 & 7.44 & 7.92 & 51.54\\
QuantizedTran & 10.40 & 13.35 & 11.07 & 80.09 & 10.44 & 17.67 & 12.60 & 104.66 & 10.58 & 21.86 & 13.72 & 128.35\\
LexRank & 23.94 & 12.14 & 15.86 & 59.51 & 23.34 & 15.96 & 18.57 & 77.33 & 23.47 & 20.03 & 21.19 & 94.43\\
TextRank & 30.45 & 15.37 & 20.10 & 73.35 & 28.18 & 18.92 & 22.24 & 92.46 & 27.00 & 22.59 & 24.17 & 110.42\\
\hdashline
\textbf{StreamHover} & \textbf{36.18} & \textbf{18.21} & \textbf{23.87} & \textbf{88.04} & \textbf{34.86} & \textbf{23.29} & \textbf{27.52} & \textbf{113.40} & \textbf{33.92} & \textbf{28.42} & \textbf{30.47} & \textbf{137.02}\\
\hline
\end{tabular}
\end{footnotesize}
\vspace{-0.05in}
\caption{Classification performance of extractive summarizers on identifying ground-truth summary utterances.}
\label{tab:classification}
\vspace{-0.1in}
\end{table*}

\vspace{0.05in}
\noindent\textbf{Settings.}\quad 
We use pretrained BERT-\textsc{base} as our embedder $\texttt{\small Embed}_\theta(\cdot)$.
The model has 12 layers, 12 heads per layer and a hidden size ($H$) of 768.
A 6-layer Transformer decoder is used as the generator $\texttt{\small Generate}_\phi(\cdot)$ to reconstruct the original utterance.
The model has 8 heads per layer, a hidden size of 768, and randomly initialized parameters. 
The convolutional encoder and decoder use a kernel size of 3. 
Because our \textbf{e}mbedder is pretrained and the \textbf{r}emaining parameters are not, we divide them into two groups $\mathcal{E}$=$\{\theta\}$ and $\mathcal{R}$=$\{\phi, \varphi\}$, then apply separate training schedules.
Following Liu and Lapata~\shortcite{liu-lapata-2019-text} we use two Adam optimizers:
\begin{align*}
lr_{\mathcal{E}} &= \widetilde{lr}_{\mathcal{E}} \cdot \texttt{\small min}(\mbox{step}^{-0.5}, \mbox{step} \cdot \mbox{warmup}_{\mathcal{E}}^{-1.5}),\\
lr_{\mathcal{R}} &= \widetilde{lr}_{\mathcal{R}} \cdot \texttt{\small min}(\mbox{step}^{-0.5}, \mbox{step} \cdot \mbox{warmup}_{\mathcal{R}}^{-1.5})
\end{align*}
where the learning rate for the embedder $\widetilde{lr}_{\mathcal{E}}$=$7e^{-4}$ is smaller than that of the rest params $\widetilde{lr}_{\mathcal{R}}$=$4e^{-2}$.
Its warmup period is longer: $\mbox{warmup}_{\mathcal{E}}$=3,000 for the embedder and $\mbox{warmup}_{\mathcal{R}}$=1,500 for the rest. 
It allows the pretrained embedder to be updated in a slower pace until other model parameters start to generate accurate gradients. 

All of our models are trained for 30 epochs on dual NVIDIA V100 GPUs with gradient accumulation every ten steps. 
We experiment with different numbers of filters, $D=\{64, \underline{100}, 128\}$, for the convolutional encoder and decoder.
The number of latent codes are varied in $K=\{512, \underline{1024}, 2048\}$. 
The coefficient $\beta$ used for commitment loss is set to 0.25 (Eq.~(\ref{eq:loss})).
These hyperparameters are tuned on the validation set. 
We keep only utterances that contain >5 words in consideration. 
The final training set contains 168,111 utterances.


\begin{table}
\setlength{\tabcolsep}{5.2pt}
\renewcommand{\arraystretch}{1.15}
\centering
\begin{footnotesize}
\begin{tabular}{|l|l||rrr|r|}
\hline
& \textbf{System} & \textbf{R-1} & \textbf{R-2} & \textbf{R-L} & \textbf{\#Wrds}\\
\hline
\hline
\multirow{4}{*}{\rotatebox[origin=c]{90}{\textbf{Abstract}}} & BART-large & 22.98 & 8.35 & 15.04 & 123.05\\
& BART-large-cnn & 23.03 & 8.03 & 16.62 & 43.13\\
& T5-large & 24.20 & 8.55 & 17.56 & 50.98\\
& FluCovRank & 25.29 & 10.93 & 16.28 & 50.00\\
\hline
\hline
\multirow{6}{*}{\rotatebox[origin=c]{90}{\textbf{Extract}}} & LEAD-5 & 24.65 & 9.54 & 17.59 & 64.04\\
& SumBasic & 23.15 & 5.57 & 14.76 & 51.54\\
& QuantizedTran & 23.90 & 7.90 & 15.37 & 128.35\\
& LexRank & \textbf{26.14} & 10.18 & 18.24 & 94.43\\
& TextRank & 25.94 & 11.44 & 18.89 & 110.42\\
\cdashline{2-6}
& \textbf{StreamHover} & 25.62 & \textbf{12.70} & \textbf{19.33} & \textbf{137.02}\\
& Oracle (Extract) & 43.42 & 30.58 & 37.99 & 110.51\\
\hline
\end{tabular}
\end{footnotesize}
\vspace{-0.05in}
\caption{Results of extractive and abstractive summarizers evaluated by ROUGE.
Extractive summarizers generate a 5-utterance summary for each clip. 
\emph{Oracle} contains ground-truth summary utterances. 
StreamHover achieves the highest scores on R-2 and R-L.
}
\label{tab:rouge_scores}
\vspace{-0.05in}
\end{table}

\begin{table}
\setlength{\tabcolsep}{6.2pt}
\renewcommand{\arraystretch}{1.15}
\centering
\begin{footnotesize}
\begin{tabular}{|l|l|ccc|}
\hline
& \textbf{System} & \textbf{Fluency} & \textbf{Informa.} & \textbf{Overall}\\
\hline
\hline
\multirow{4}{*}{\rotatebox[origin=c]{90}{\textbf{Human}}} & FluCovRank & -0.95 & -0.93 & -0.97 \\
& LexRank & 0.25 & 0.11 & 0.17 \\
& BART-large & 0.28 & 0.31 & 0.28\\
\cdashline{2-5}
& \textbf{StreamHover} & \textbf{0.42} & \textbf{0.52} & \textbf{0.52} \\
\hline
\end{tabular}
\end{footnotesize}
\vspace{-0.05in}
\caption{Results of human evaluation regarding fluency, informativeness and the overall quality of system summaries using Best-Worst Scaling. 
}
\label{tab:human_eval}
\vspace{-0.1in}
\end{table}

\begin{table}[t]
\setlength{\tabcolsep}{5pt}
\renewcommand{\arraystretch}{1.08}
\begin{scriptsize}

\begin{minipage}[b]{0.5\hsize}\centering
\begin{tabular}[t]{|p{2.88in}|}


\hline

\textbf{FluCovRank}\\[1mm]

\textbullet\, top left bottom 
/ cloud studies today 
/ find links to their original posts 
/ hey jennifer saw the images
/ love the top left and bottom
/ info tab and i uploaded
/ colors are beautiful but im partial through colorful sky scenes
/ pretty large about 4000 by 4000 pixels
/ photo studies of today
/ moment\\[1mm]

\hline
\hline

\textbf{LexRank}\\[1mm]

\textbullet\, I hope you guys are having a good day so far.\\[0.5mm]

\textbullet\, So I'm going to be \underline{painting from these images} and these beautiful photos are from various photographers.\\[0.5mm]

\textbullet\, Those yeah well \underline{top right also is like very Contra high contrast that tends} \underline{to like grab my attention} when I look at the sheet but I would say \underline{top left} \underline{and bottom right give me the most like happy feels.}\\[0.5mm]

\textbullet\, So yeah, if you guys want to \underline{grab the reference images}, you can find them in the stream description below the individual images...\\[0.5mm]


\hline
\hline

\textbf{BART-Large}\\[1mm]

\textbullet\, Hello good morning everybody welcome to the stream. I hope you guys are having a good day so far. Is there a lot of buffering or are we doing alright? I got a little message that there was some connectivity issue. For a moment there, so I hope I hope it's OK. Yeah, I'll just keep going. So yeah, if you guys want to \underline{grab the reference images}, you can find them in the stream description below the individual images... \\[1mm]

\hline
\hline

\textbf{Quantized Transformer}\\[1mm]

\textbullet\, Good to see you were going to be doing cloud studies today.\\[0.1mm]
\textbullet\, The stream in the description.\\[0.1mm]
\textbullet\, \underline{One of them is from Morguefile}, One is from unsplash, well, \underline{two are} \underline{from Unsplash} and \underline{one is from pixels} there a little bit from all over the place, but you can find the photographers below if you'd like.\\[0.1mm]
\textbullet\, Hey Jennifer, saw the images.\\[0.1mm]
\textbullet\, Let's see top left, bottom right...\\[0.1mm]

\hline
\hline

\textbf{StreamHover (Ours)}\\[1mm]

\textbullet\, So if anybody is interested in joining in, if you want to \underline{work on some} \underline{skies for your landscapes} for future landscapes, this is what we're going to be doing.\\[1mm]

\textbullet\, \underline{One of them is from Morguefile}, One is from unsplash, well, \underline{two are} \underline{from Unsplash} and \underline{one is from pixels} there a little bit from all over the place, but you can find the photographers below if you'd like.\\[1mm]

\textbullet\, Those yeah well \underline{top right also is like very Contra high contrast that tends} \underline{to like grab my attention} when I look at the sheet but I would say \underline{top left} \underline{and bottom right give me the most like happy feels.}\\[1mm]

\textbullet\, So yeah, if you guys want to \underline{grab the reference images}, you can find them in the stream description below the individual images...\\[1mm]


\hline
\end{tabular}
\end{minipage}

\end{scriptsize}
\vspace{-0.05in}
\caption{
Example system summaries for \emph{Digital Painting Studies with Maddy Bellwoar--Clouds}.
The BART summary is fluent but its content lacks specificity, as is the case for LexRank.
The summary segments selected by FluCovRank are ungrammatical. 
StreamHover identifies on-topic and informative utterances related to digital painting.
Their relevant spans of text are manually underlined for readability.
}
\label{tab:results_output}
\vspace{-0.15in}
\end{table}

\begin{table}[t]
\setlength{\tabcolsep}{1.8pt}
\renewcommand{\arraystretch}{1.1}
\centering
\begin{scriptsize}
\begin{tabular}{|rp{2.3in}rrr|}
\hline
& \textbf{Utterances} & \textbf{C1} & \textbf{C2} & \textbf{C3}\\
\hline
\hline
0 & Hello good morning everybody welcome high foster highly art. & $\square$ & $\square$ & \done\\
1 & \st{Hi Lisa, welcome everyone.} & $\square$ & $\square$ & $\square$\\
2 & I hope you guys are having a good day so far. & $\square$ & $\square$ & $\square$\\
3 & Good to see you were going to be doing cloud studies today. & $\square$ & $\square$ & $\square$\\
4 & {\cellcolor[gray]{.88}}So if anybody is interested in joining in, if you want to work on some skies for your landscapes for future landscapes, this is what we're going to be doing. & $\square$ & \done & \done\\
5 & \st{Photo studies of today.} & $\square$ & $\square$ & $\square$\\
6 & {\cellcolor[gray]{.88}}So I'm going to be painting from these images and these beautiful photos are from various photographers. & $\square$ & $\square$ & \done\\
7 & You can find links to their original posts below. & $\square$ & $\square$ & $\square$\\
8 & The stream in the description. & \done & $\square$ & $\square$\\
9 & One of them is from Morguefile, One is from unsplash, well, two are from Unsplash and one is from pixels there a little bit from all over the place, but you can find the photographers below if you'd like. & $\square$ & \done & $\square$\\
10 & Hey Jennifer, saw the images. & $\square$ & $\square$ & $\square$\\
11 & I really love the top left and bottom right. & $\square$ & $\square$ & $\square$\\
12 & The colors are beautiful but I'm partial through colorful Sky scenes. & $\square$ & $\square$ & $\square$\\
13 & \st{Yeah, I totally agree.} & $\square$ & $\square$ & $\square$\\
14 & Let's see top left, bottom right. & $\square$ & $\square$ & $\square$\\
15 & {\cellcolor[gray]{.88}}Those yeah well top right also is like very Contra high contrast that tends to like grab my attention when I look at the sheet but I would say top left and bottom right give me the most like happy feels. & $\square$ & \done & $\square$\\
& ... & & & \\
\hline
\end{tabular}
\end{scriptsize}
\vspace{-0.05in}
\caption{
A snippet from \emph{Digital Painting Studies with Maddy Bellwoar--Clouds}.
We show the most prominent latent codes and their representative utterances (`X').
Human annotated summary utterances are colored gray and ultra-short utterances are crossed out.
}
\label{tab:code}
\vspace{-0.15in}
\end{table}

\subsection{Results}
\label{sec:results}

In Table~\ref{tab:classification}, we analyze the performance of extractive summarizers on identifying ground-truth summary utterances and report their precision, recall and $F_1$-measure scores. 
We vary the length of their output to yield \{3, 4, 5\}-utterance summaries. 
In comparison, a ground-truth extract contains 5.5 utterances.
The Lead-N baseline selects the first N utterances of a clip.
It gives low scores because our data do not present strong lead bias as that of news articles.
We find that StreamHover consistently outperforms other summarization systems across all lengths.
Its length, when measured by number of words, is comparable to that of LexRank and TextRank.
The highest $F_1$-score of 30.47\% is achieved when StreamHover generates a 5-utterance summary for each 5-minute clip.
This amounts to rendering one utterance per one-minute segment when a user scrolls past the video.

In Table~\ref{tab:rouge_scores}, we compare extractive and abstractive summarizers and report ROUGE scores~\cite{lin-2004-rouge} that measure content overlap between system and reference summaries.\footnote{
The recent automatic metrics~\cite{zhang2020bertscore,sellam-etal-2020-bleurt} have not been tested on speech transcripts. 
Spoken text contains filled pauses (um, uh, well), disfluencies (go-go-go away), repetitions and verbal interruptions. 
ROUGE is the only metric that has been validated to attain good correlation with human judgments on transcripts~\cite{liu-liu-2008-correlation}.
}
We use human abstracts as the reference.
All extractive summarizers produce 5-utterance summaries and \emph{Oracle Extract} contains ground-truth utterances. 
It places an upper bound on the performance of extractive summarizers.
We observe that StreamHover yields the highest scores on both R-2 and R-L metrics. 

\begin{figure}[t]
\centering
\includegraphics[width=2.9in]{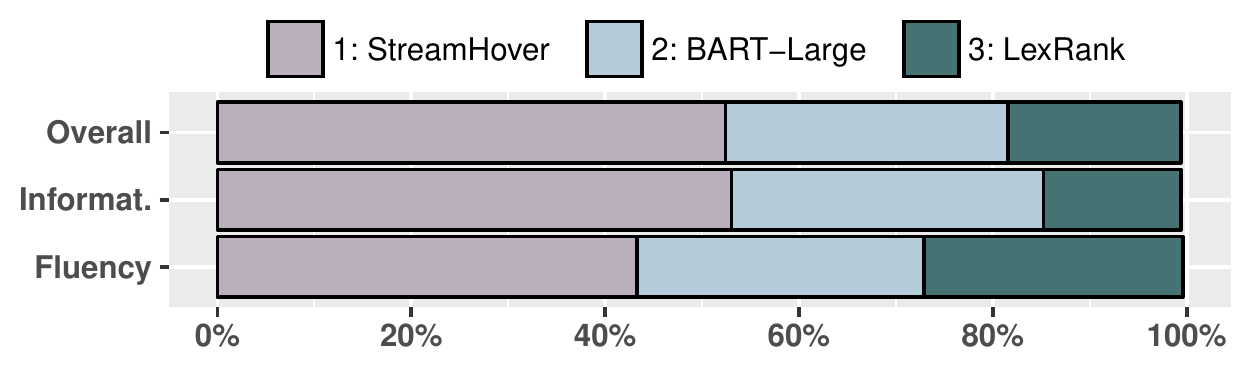}
\vspace{-0.1in}
\caption{The proportion of times a system is selected as the ``Best.'' FluCovRank is omitted as it is <1\%.}
\label{fig:best-percent}
\vspace{-0.15in}
\end{figure}

We show example system summaries in Table~\ref{tab:results_output}.
The abstractive summaries generated by BART-Large are fluent but their content lacks specificity, 
so are the summaries produced by LexRank and Quantized Transformer.
Particularly, QT does not seem to perform well on this task despite that the model has been retrained on livestream transcripts.\footnote{https://github.com/stangelid/qt/blob/main/custom.md}
We believe this is partly because words and phrases tend to repeat themselves in review documents, 
and while spoken utterances are verbose, there is little repetition found in the transcripts.
We observe that summary segments selected by FluCovRank are on-topic but they are ungrammatical and difficult to interpret without context. 
In contrast, StreamHover can identify on-topic and informative utterances related to digital painting.
We provide more examples in the supplementary materials.

In Table~\ref{tab:code}, we study the most prominent latent codes (C1-3) and their associated utterances.
We define representative utterances as those frequently assigned to these codes (Eq. (\ref{eq:z_i})).
We observe that C1 usually contains a skewed number of utterances that are commonly seen in the data and not representative of the input;
C2 contains lengthy but not necessarily summary-worthy utterances. 
In our experiments, we exclude C1/C2 before performing grid search on all codes to find the set of prominent codes:
we use $\mathcal{P}$=50 tuned on the valid set which is effective in helping identify summary utterances.\footnote{
For increased training stability of variational autoencoder (VAE) models, we refer the reader to~\cite{li-etal-2019-surprisingly}.
}

We conduct a human evaluation to assess how StreamHover compares to strong extractive and abstractive baselines. 
They are (a) LexRank~\cite{Erkan:2004}, (b) FluCovRank~\cite{shang-etal-2018-unsupervised} and (c) BART-Large~\cite{lewis-etal-2020-bart};
the latter two are abstractive systems.
Each evaluator is shown a video clip with a synchronized display of the transcript followed by four system summaries,
shown in random order to remove any positional bias.
The evaluator is asked to select the best and worst of the summaries according to each of these criteria: 
\emph{Fluency/Coherence}: is the the summary well-presented, grammatically correct and easy to read? 
\emph{Informativeness}: does the summary provide useful information about the video clip? 
\emph{Overall Quality}: is the summary of good quality considering both content and linguistic aspects?

We randomly sample 100 clips from the test set.
Each clip and its summaries are judged by five evaluators that we recruit from Amazon mechanical turk.\footnote{
They are required to be Master workers, have an approval rate >90\% and completed >1K tasks, and located in the U.S.}
Table~\ref{tab:human_eval} shows the performance of all systems measured by Best-Worst Scaling~\cite{kiritchenko-mohammad-2016-capturing}, where the score of a system is computed as the percentage of times it was selected as the best minus the worst. The range of scores is [-1,1].
Figure~\ref{fig:best-percent} shows how frequently a system is chosen to produce the ``best summary.'' 
We observe that StreamHover achieves an overall score of 0.52 and it is selected as the best summary in over half of the times.






\section{Conclusion}
\label{sec:conclusion}

We present StreamHover, a new framework for annotating and summarizing livestream transcripts.
Our dataset contains over 500 hours of videos annotated with extractive and abstractive summaries.
We explored an extractive method leveraging VQ-VAE to identify salient summary utterances and obtained strong results.
Future work includes boosting summarization solutions to provide users a concentrated overview of streamed content. 


\bibliographystyle{acl_natbib}
\bibliography{summ,more,anthology}

\clearpage
\setcounter{page}{1}

\appendix
\section{B\={e}hance Dataset}
\label{app:datagathering}

We collect a total of 5,398 streamed videos from \textsf{\small Behance.net}. 
Some streamers opt-out of the transcription service provided by Microsoft Automatic Speech Recognition, so transcripts are not available for these videos.
We create a list of domain keywords by finding 50 most frequently appearing words from video titles (stopwords are excluded).
Examples include `fresco', `adobe', `photoshop', `illustration', `art', `painting', `drawing', `illustrator', `character', `design.' 
The keywords are used to select videos for human annotation. 
2,360 videos have transcripts available and contain at least one of our domain keywords in their titles. 
These videos are split into clips of 5-minute each.  
Some clips contain little or no verbal content. 
We thus remove clips that contain very few words ($\leq$333 words) or utterances ($\leq$38 utterances).
These thresholds are determined using the average values of all clips. 
Videos with less than 5 valid clips are also removed from consideration. 
This preprocessing step gives 6,003 clips from 381 videos.
During annotation, our annotators find 582 clips to contain only chit-chats, suggesting that these clips are uninformative. 
11 videos contain only chit-chat clips, they are subsequently removed from the dataset, yielding a total of 5,421 clips from 370 videos that are split into train, validation and test sets.

\section{Baseline Summarizers}

Our neural abstractive baselines include pre-trained BART-large~\cite{lewis-etal-2020-bart}, BART-large-cnn, and T5-large~\cite{JMLR:v21:20-074}.
We follow the HuggingFace implementation~\cite{wolf-etal:2020-transformers}. 
Utterances that are longer than 5 words are concatenated into a flat sequence, which is used as the input to each summarizer.
The model parameters include: the maximum and minimum summary lengths are 150 and 15 tokens, respectively.
We use a beam size of 5 with early stopping.
The length penalty is 1.0. 
``no\_repeat\_ngram\_size'' is set to 3, such that a trigram cannot occur more than once in the summary.

Our extractive baselines include LexRank~\cite{Erkan:2004}, TextRank~\cite{mihalcea-tarau-2004-textrank}, and SumBasic~\cite{Vanderwende:2007}.
They are implemented using the Sumy library where we adopt the default text parser and stemmer.
Our unsupervised summarizer for speech transcript summarization~\cite{shang-etal-2018-unsupervised} uses the following settings: we report the FluCovRank scores. The number of components used in LSA is 25. The number of utterance communities is 35.
The number of clusters is 6, with a scaling factor of 1.3 and lambda of 0.4. The size of the summary is set to 50 words.

\section{Example Summaries}

We show example summaries generated by different summarizers: FluConvRank~\cite{shang-etal-2018-unsupervised}, LexRank~\cite{Erkan:2004}, BART-large~\cite{lewis-etal-2020-bart} and StreamHover.
We also show the top-3 most prominent latent codes and their associated utterances.
We choose 5 representative utterances for each code that are most frequently assigned to this code. 
We observe that C1 utterances are frequently seen in the data (chit-chats) and not representative of the input.
C2 is associated with lengthy but not necessarily summary-worthy utterances. 
C3 utterances are both comprehensive and contain diverse information.
In our experiments, we exclude C1/C2 before performing grid search on all codes to find the set of prominent codes.
It allows us to effectively identify summary utterances without biasing towards the lengthy ones.

\clearpage

\begin{table}[H]
\setlength{\tabcolsep}{5pt}
\renewcommand{\arraystretch}{1.1}
\begin{scriptsize}

\begin{minipage}[b]{0.5\hsize}\centering
\begin{tabular}[t]{|p{2.99in}|}


\hline

\textbf{FluCovRank}\\[1mm]

\textbullet\, digital painting today / 3 hours minimum / explain my painting process / starbucks florida cup / really really cheerful scene / prob hey welcome hi kumar / canvas sizes 3000 by 2000 pixels / sore throat so todays stream might be a tiny bit shorter / otherwise just sit back and relax and enjoy the painting / find \\[1mm]

\hline
\hline

\textbf{LexRank}\\[1mm]

\textbullet\, If you guys are interested in painting along with me or doing a little sketch or whatever, you have time for your always welcome to my reference.\\[0.5mm]

\textbullet\, Is what I have planned to work on this if I'm feeling good, we might go on or do another painting, but I'm drinking some tea with honey and I have to kind of be careful with my voice a little bit talking so much, but I'm otherwise I'm good.\\[0.5mm]

\textbullet\, So again, that's where you can find my references below in the stream description.\\[0.5mm]

\textbullet\, You guys are welcome to paint along if you're interested in doing that.\\[0.5mm]

\textbullet\, I've lived in a few different states, but where I spent most of my life was in Florida so.\\[1mm]

\hline
\hline

\textbf{BART-Large}\\[1mm]

\textbullet\, Good afternoon, good evening, good morning, good day, good afternoon, how are you? Good morning, hi Mohammed, hi kumar, hi muhammad, hi hi kumar, hi hello hi hi hi Mohammed. I hope you've had a good day so far. I'm doing pretty good pro to be honest. I have a little bit of a sore throat, so today's stream might be a tiny bit shorter than some of the others. It's still going to be around 3 hours minimum. Is what I have planned to work on this if I'm feeling good, we might go on or do another painting, but I'm drinking some tea with honey and I have to kind of\\[1mm]

\hline
\hline

\textbf{Quantized Transformer}\\[1mm]

\textbullet\,  We're going to be doing some digital painting today.\\[0.1mm]

\textbullet\,  As usual, I'm going to be working from this photo as a reference.\\[0.1mm]

\textbullet\,  I hope you're all doing well.\\[0.1mm]

\textbullet\,  Image is below in the stream description and above the chat.\\[0.1mm]

\textbullet\,  It's still going to be around 3 hours minimum.\\[0.1mm]

\textbullet\,  Is what I have planned to work on this if I'm feeling good, we might go on or do another painting, but I'm drinking some tea with honey and I have to kind of be careful with my voice a little bit talking so much, but I'm otherwise I'm good.\\[1mm]

\textbullet\,  Um prob hey welcome Hi Kumar.\\[0.1mm]

\textbullet\,  You guys are welcome to paint along if you're interested in doing that.\\[0.1mm]

\textbullet\,  I hope it's going to be good.\\[0.1mm]

\textbullet\,  My canvas sizes 3000 by 2000 pixels.\\[1mm]

\hline
\hline

\textbf{StreamHover (Ours)}\\[1mm]

\textbullet\, It's like a little cafe table in Spain, with some dappled lighting coming through these trees.\\[1mm]

\textbullet\, I really like the pattern of light on the building over here, and the just overall feeling the different colors.\\[1mm]

\textbullet\, If you guys are interested in painting along with me or doing a little sketch or whatever, you have time for your always welcome to my reference.\\[1mm]

\textbullet\, Is what I have planned to work on this if I'm feeling good, we might go on or do another painting, but I'm drinking some tea with honey and I have to kind of be careful with my voice a little bit talking so much, but I'm otherwise I'm good.\\[1mm]

\textbullet\, I am using pure ref to put the little reference image up here so we can see a thumbnail view of it while we're painting.\\[1mm]

\hline
\end{tabular}
\end{minipage}

\end{scriptsize}
\caption{
Example system summaries from \emph{Digital Painting with Maddy Bellwoar}.
BART summary is fluent but its content lacks specificity, as is the case for LexRank.
Summary segments selected by FluCovRank are ungrammatical. 
StreamHover identifies on-topic and informative utterances related to digital painting.
}
\label{tab:results_output1}
\vspace{-0.10in}
\end{table}

\begin{table}[H]
\setlength{\tabcolsep}{1.8pt}
\renewcommand{\arraystretch}{1.1}
\centering
\begin{scriptsize}
\begin{tabular}{|rp{2.3in}rrr|}
\hline
& \textbf{Utterances} & \textbf{C1} & \textbf{C2} & \textbf{C3}\\
\hline
\hline
0 & \st{Hi everybody, good morning.} & $\square$ & $\square$ & $\square$\\
1 & Good afternoon, good evening whatever time it is for you. & $\square$ & $\square$ & $\square$\\
2 & \st{Thanks for being here.} & $\square$ & $\square$ & $\square$\\
3 & \st{Hi Fernando Allison, Welcome.} & $\square$ & $\square$ & $\square$\\
4 & I hope you've had a good day so far. & $\square$ & $\square$ & $\square$\\
5 & I hope you've all been well. & $\square$ & $\square$ & $\square$\\
6 & {\cellcolor[gray]{.88}}If I haven't seen you since last week, I hope you had a good weekend. & $\square$ & $\square$ & $\square$\\
7 & We're going to be doing some digital painting today. & $\square$ & $\square$ & $\square$\\
8 & As usual, I'm going to be working from this photo as a reference. & $\square$ & $\square$ & $\square$\\
9 & I think it's really beautiful. & \done & $\square$ & $\square$\\
10 & It's like a little cafe table in Spain, with some dappled lighting coming through these trees. & $\square$ & \done & \done\\
11 & {\cellcolor[gray]{.88}}I really like the pattern of light on the building over here, and the just overall feeling the different colors. & $\square$ & \done & \done\\
12 & the Blues and pinks and yellows on the wall here. & $\square$ & $\square$ & $\square$\\
13 & It's really, really cheerful scene. & $\square$ & $\square$ & $\square$\\
14 & Hi Mohammed, high pro, how are you? & $\square$ & $\square$ & $\square$\\
15 & I hope you're all doing well. & \done & $\square$ & $\square$\\
16 & If you guys are interested in painting along with me or doing a little sketch or whatever, you have time for your always welcome to my reference. & $\square$ & \done & \done\\
17 & Image is below in the stream description and above the chat. & $\square$ & $\square$ & $\square$\\
18 & Also, if you check the info panel it should be there as well so this photo is from Unsplash. & $\square$ & $\square$ & \done\\
19 & You can find a link to the original post from the photographer below. & $\square$ & $\square$ & $\square$\\
20 & \st{The stream.} & $\square$ & $\square$ & $\square$\\
21 & I'm doing pretty good pro to be honest. & $\square$ & $\square$ & $\square$\\
22 & {\cellcolor[gray]{.88}}I have a little bit of a sore throat, so today's stream might be a tiny bit shorter than some of the others. & $\square$ & \done & $\square$\\
23 & It's still going to be around 3 hours minimum. & $\square$ & $\square$ & $\square$\\
24 & Is what I have planned to work on this if I'm feeling good, we might go on or do another painting, but I'm drinking some tea with honey and I have to kind of be careful with my voice a little bit talking so much, but I'm otherwise I'm good. & $\square$ & $\square$ & $\square$\\
25 & Um prob hey welcome Hi Kumar. & $\square$ & \done & $\square$\\
26 & So again, that's where you can find my references below in the stream description. & $\square$ & $\square$ & $\square$\\
27 & You guys are welcome to paint along if you're interested in doing that. & $\square$ & $\square$ & $\square$\\
28 & Otherwise, just sit back and relax and enjoy the painting. & $\square$ & $\square$ & $\square$\\
29 & I'll try to do a nice job with this one. & $\square$ & $\square$ & $\square$\\
30 & I hope it's going to be good. & \done & $\square$ & $\square$\\
31 & I really like the reference so. & \done & $\square$ & $\square$\\
32 & \st{Hi Dante.} & $\square$ & $\square$ & $\square$\\
33 & So I'm going to explain my painting process as we go through here. & $\square$ & $\square$ & $\square$\\
34 & And if you guys have any questions you can let me know. & $\square$ & $\square$ & $\square$\\
35 & {\cellcolor[gray]{.88}}I am using pure ref to put the little reference image up here so we can see a thumbnail view of it while we're painting. & $\square$ & $\square$ & \done\\
36 & My canvas sizes 3000 by 2000 pixels. & $\square$ & $\square$ & $\square$\\
37 & \st{If you're curious.} & $\square$ & $\square$ & $\square$\\
38 & Yes, I'm using a different Cup today. & $\square$ & $\square$ & $\square$\\
39 & This is my Starbucks Florida Cup. & $\square$ & $\square$ & $\square$\\
40 & Maybe more information than you need, but yeah, so I'm I grew up in Florida. & $\square$ & $\square$ & $\square$\\
41 & {\cellcolor[gray]{.88}}I've lived in a few different states, but where I spent most of my life was in Florida so. & $\square$ & $\square$ & $\square$\\
42 & I don't live there now. & \done & $\square$ & $\square$\\
\hline
\end{tabular}
\end{scriptsize}
\vspace{-0.05in}
\caption{
A transcript snippet from \emph{Digital Painting with Maddy Bellwoar}.
We show the most prominent latent codes and their representative utterances (`X').
Human annotated summary utterances are colored gray and ultra-short utterances are crossed out.
}
\label{tab:code1}
\vspace{-0.1in}
\end{table}

\begin{table}[H]
\setlength{\tabcolsep}{5pt}
\renewcommand{\arraystretch}{1.1}
\begin{scriptsize}

\begin{minipage}[b]{0.5\hsize}\centering
\begin{tabular}[t]{|p{2.99in}|}


\hline

\textbf{FluCovRank}\\[1mm]

\textbullet\, view is created by them stitching images / maps years was a fitness studio yeah / image stitching just has some weird that happens team / oh if you want to share a link for map crunch / leave so weird parts / hope achieve that feeling of the water is really moving / yeah its\\[1mm]

\hline
\hline

\textbf{LexRank}\\[1mm]

\textbullet\, So I'm going to start to work on the foreground now and this is going to be tricky, so this is one of the things that made me choose to paint this reference.\\[0.5mm]

\textbullet\, The rushing water in the foreground kind of coming towards us and the ripples that it creates an the little bubbles and all that kind of stuff.\\[0.5mm]

\textbullet\, Sometimes the image stitching just has some weird stuff that happens, Oh team.\\[0.5mm]

\textbullet\, Oh, if you want to share a link for map crunch.\\[0.5mm]

\textbullet\, Alright that should work this time.\\[1mm]

\hline
\hline

\textbf{BART-Large}\\[1mm]

\textbullet\, Trying to figure out those relationships near the end or like when you're trying to like if you're going to try to get the one or like if the one is like if I'm working on the one I'm going to do it's like if it's not working. I'm ready for some dog like with half his body has been overlapped by a boat. Yeah, it's always really satisfying the color combination in these kind of desert scenes, the nice blue Sky and like the Warm Reds and oranges. It's like you get the extra bonus of like some purple thrown in there and that's just perfect. You know, I've seen people like riding camels and stuff, he's taking pictures\\[1mm]

\hline
\hline

\textbf{Quantized Transformer}\\[1mm]

\textbullet\,  So I hope I can achieve that feeling of the water is really moving there.\\[0.1mm]

\textbullet\,  You have to click the share button and then copy the link that is there if you copy the link from the URL above it's not going to send the right image.\\[0.1mm]

\textbullet\,  So what you can do is click right here where it says share and then copy this.\\[0.1mm]

\textbullet\,  OK, so these so right now, I'm just going to start with a darker shadows.\\[0.1mm]

\textbullet\,  Of where I see the ripples.\\[0.1mm]

\textbullet\,  Alright that should work this time.\\[0.1mm]

\textbullet\,  You know where some of the really, really fantastic ones come from.\\[0.1mm]

\textbullet\,  But people will be on the boat on foot.\\[0.1mm]

\textbullet\,  Maps years was a fitness studio, yeah, there's all kinds of stuff.\\[0.1mm]

\textbullet\,  I've painted with references with stuff like that.\\[1mm]

\hline
\hline

\textbf{StreamHover (Ours)}\\[1mm]

\textbullet\, So I'm going to start to work on the foreground now and this is going to be tricky, so this is one of the things that made me choose to paint this reference.\\[1mm]

\textbullet\, The rushing water in the foreground kind of coming towards us and the ripples that it creates an the little bubbles and all that kind of stuff.\\[1mm]

\textbullet\, Yeah, it's always really satisfying the color combination in these kind of desert scenes, the nice blue Sky and like the Warm Reds and oranges.\\[1mm]

\textbullet\, I'm ready for some dog like with half his body has been overlapped by a boat.\\[1mm]

\textbullet\, People are in a car traveling there, you get a lot of side of the road like kind of bland images, but then you also get like the most amazing pictures and people also upload images and stuff, I think that might be.\\[1mm]

\hline
\end{tabular}
\end{minipage}

\end{scriptsize}
\caption{
Example system summaries from \emph{Virtual Plein Air Landscape Painting}.
The BART summary is fluent but its content lacks specificity, as is the case for LexRank.
The summary segments selected by FluCovRank are ungrammatical. 
StreamHover identifies on-topic and informative summary utterances.
}
\label{tab:results_output2}
\vspace{-0.1in}
\end{table}

\begin{table}[H]
\setlength{\tabcolsep}{1.8pt}
\renewcommand{\arraystretch}{1.1}
\centering
\begin{scriptsize}
\begin{tabular}{|rp{2.3in}rrr|}
\hline
& \textbf{Utterances} & \textbf{C1} & \textbf{C2} & \textbf{C3}\\
\hline
\hline
0 & Trying to figure out those relationships near the end or like force them if they're not working. & $\square$ & $\square$ & $\square$\\
1 & {\cellcolor[gray]{.88}}So I'm going to start to work on the foreground now and this is going to be tricky, so this is one of the things that made me choose to paint this reference. & $\square$ & $\square$ & \done\\
2 & 'cause I wanted to figure out how to paint this kind of situation. & $\square$ & $\square$ & $\square$\\
3 & The rushing water in the foreground kind of coming towards us and the ripples that it creates an the little bubbles and all that kind of stuff. & $\square$ & \done & \done\\
4 & So I hope I can achieve that feeling of the water is really moving there. & $\square$ & $\square$ & $\square$\\
5 & Might give some inspiration alright let's see got an artstation link. & $\square$ & $\square$ & $\square$\\
6 & \st{Oh, cool.} & $\square$ & $\square$ & $\square$\\
7 & \st{Good vibes good vibes.} & $\square$ & $\square$ & $\square$\\
8 & \st{Other desert desert scene.} & $\square$ & $\square$ & $\square$\\
9 & Yeah, it's always really satisfying the color combination in these kind of desert scenes, the nice blue Sky and like the Warm Reds and oranges. & $\square$ & \done & \done\\
10 & I like the one I'm working on 2 'cause. & $\square$ & $\square$ & $\square$\\
11 & {\cellcolor[gray]{.88}}It's like you get the extra bonus of like some purple thrown in there and that's just perfect. & $\square$ & $\square$ & \done\\
12 & \st{For me.} & $\square$ & $\square$ & $\square$\\
13 & \st{Yeah, that's cool thanks.} & $\square$ & $\square$ & $\square$\\
14 & {\cellcolor[gray]{.88}}Team I says, I was looking for a nice location on map French and I found this alright. & $\square$ & $\square$ & $\square$\\
15 & I'm ready for some dog like with half his body has been overlapped by a boat. & $\square$ & $\square$ & $\square$\\
16 & Sometimes the image stitching just has some weird stuff that happens, Oh team. & $\square$ & $\square$ & $\square$\\
17 & Oh, if you want to share a link for map crunch. & $\square$ & $\square$ & $\square$\\
18 & {\cellcolor[gray]{.88}}You have to click the share button and then copy the link that is there if you copy the link from the URL above it's not going to send the right image. & $\square$ & \done & $\square$\\
19 & So what you can do is click right here where it says share and then copy this. & $\square$ & $\square$ & $\square$\\
20 & And then that will be the right one, 'cause I think it sent the wrong one from you. & $\square$ & $\square$ & $\square$\\
21 & \st{Yeah.} & $\square$ & $\square$ & $\square$\\
22 & \st{Alright cool.} & $\square$ & $\square$ & $\square$\\
23 & Will get back to painting? & \done & $\square$ & $\square$\\
24 & OK, so these so right now, I'm just going to start with a darker shadows. & $\square$ & $\square$ & $\square$\\
25 & Of where I see the ripples. & \done & $\square$ & $\square$\\
26 & I'm not worried about putting like every ripple in the right spot. & $\square$ & $\square$ & $\square$\\
27 & But we couldn't get the overall feel this. & $\square$ & $\square$ & $\square$\\
28 & \st{Some more Brown here.} & $\square$ & $\square$ & $\square$\\
29 & Alright that should work this time. & \done & $\square$ & $\square$\\
30 & What's wrong with that picture? & \done & $\square$ & $\square$\\
31 & \st{Does actually seems nice?} & $\square$ & $\square$ & $\square$\\
32 & I don't see anything wrong with that landscape Timo? & $\square$ & $\square$ & $\square$\\
33 & When you when you're using map crunch you're going to get a lot of like side of the road road views because you know if you think about where the car is going. & $\square$ & \done & $\square$\\
34 & Order fact that it's your in the Google Maps. & $\square$ & $\square$ & $\square$\\
35 & {\cellcolor[gray]{.88}}People are in a car traveling there, you get a lot of side of the road like kind of bland images, but then you also get like the most amazing pictures and people also upload images and stuff, I think that might be. & $\square$ & \done & \done\\
36 & You know where some of the really, really fantastic ones come from. & $\square$ & $\square$ & $\square$\\
37 & But people will be on the boat on foot. & $\square$ & $\square$ & $\square$\\
38 & You know, I've seen people like riding camels and stuff, he's taking pictures for Google. & $\square$ & $\square$ & $\square$\\
39 & Maps years was a fitness studio, yeah, there's all kinds of stuff. & $\square$ & $\square$ & $\square$\\
40 & {\cellcolor[gray]{.88}}Yeah, but sometimes you get something weird because the 3D or the 360 Panorama. & $\square$ & $\square$ & $\square$\\
41 & View is created by them stitching images together. & $\square$ & $\square$ & $\square$\\
42 & So, sometimes you have weird situations where stuff overlaps. & $\square$ & $\square$ & $\square$\\
\hline
\end{tabular}
\end{scriptsize}
\vspace{-0.05in}
\caption{A snippet from \emph{Virtual Plein Air Landscape Painting}.
We show the most prominent latent codes and their representative utterances (`X').
Human annotated summary utterances are colored gray and ultra-short utterances are crossed out.
}
\label{tab:code2}
\vspace{-0.1in}
\end{table}

\begin{table}[H]
\setlength{\tabcolsep}{5pt}
\renewcommand{\arraystretch}{1.1}
\begin{scriptsize}

\begin{minipage}[b]{0.5\hsize}\centering
\begin{tabular}[t]{|p{2.99in}|}


\hline

\textbf{FluCovRank}\\[1mm]

\textbullet\, drawing diesel letters / surface pen 4 i have the 3 / latest song version of the surface / pin so i got a surface pen / uhm this pan to check / avoid these artifacts in astana / file ready an complete im just listening / early in october 20th 19th / bump on phone so yeah\\[1mm]

\hline
\hline

\textbf{LexRank}\\[1mm]

\textbullet\, Thank you for sticking around wanna show you some real cool stuff, not right there.\\[0.5mm]

\textbullet\, Bump on Phone so yeah, that's the wrong letter, but let's go.\\[0.5mm]

\textbullet\, But I thought we fresco does not have good, um Palm rejection and so when you using the pen and drawing whether you're on the iPad or surface.\\[0.5mm]

\textbullet\, We go come on come here pin so I got a Surface Pen here.\\[0.5mm]

\textbullet\, Pan's had details that were missing right now, I'm using the vector pen the vector brushes are in.\\[1mm]

\hline
\hline

\textbf{BART-Large}\\[1mm]

\textbullet\, Thank you for sticking around wanna show you some real cool stuff, not right there. I'm on the Surface Pro. That's the latest song version of the surface. It just came out early in October 20th 19th. The app and it's working it's a work in progress OK is constantly being updated with the features and bugs. They sent me uhm this pan to check out. Pan's had details that were missing right now, I'm using the vector pen the vector brushes are in. So I'm doing this little dots and what happens is we on with Adobe fresco is there. But I thought we fresco does not have good, um Palm rejection and so when you using the\\[1mm]

\hline
\hline

\textbf{Quantized Transformer}\\[1mm]

\textbullet\,  I do have my Surface Pen somewhere it's up there.\\[0.1mm]

\textbullet\,  I'm on the Surface Pro.\\[0.1mm]

\textbullet\,  That's the latest song version of the surface.\\[0.1mm]

\textbullet\,  You end up with all these little artifacts that when your Palm lays on the screen.\\[0.1mm]

\textbullet\,  And so as I was drawing diesel letters.\\[0.1mm]

\textbullet\,  I got to the point where.\\[0.1mm]

\textbullet\,  I just cannot avoid these artifacts in Astana kept them.\\[0.1mm]

\textbullet\,  I need to do some of these lines here is geared is closeout all.\\[0.1mm]

\textbullet\,  If this doesn't have a pain.\\[0.1mm]

\textbullet\,  Just did good thing is I have extra pants, not only do I have all here's one, not only do I have extra pants.\\[0.1mm]

\textbullet\,  OK, here, we go ever it is yeah, some of these.\\[1mm]

\hline
\hline

\textbf{StreamHover (Ours)}\\[1mm]

\textbullet\, It's gotta stay and the same letter so they was the letter C and I'm just using my hands my fingers.\\[1mm]

\textbullet\, The app and it's working it's a work in progress OK is constantly being updated with the features and bugs.\\[1mm]

\textbullet\, But I thought we fresco does not have good, um Palm rejection and so when you using the pen and drawing whether you're on the iPad or surface.\\[1mm]

\textbullet\, I kept them in the drawing and now they are part of the look of each of these letters, which is crazy.\\[1mm]

\textbullet\, Just did good thing is I have extra pants, not only do I have all here's one, not only do I have extra pants.\\[1mm]

\hline
\end{tabular}
\end{minipage}

\end{scriptsize}
\caption{
Example system summaries from \emph{Creating ABC Childrens Book Art on Adobe Fresco and Adobe Illustrator Part 18}.
BART summary is fluent but its content lacks specificity, as is the case for LexRank.
Summary segments selected by FluCovRank are ungrammatical. 
StreamHover identifies on-topic and informative summary utterances.
}
\label{tab:results_output3}
\vspace{-0.1in}
\end{table}

\begin{table}[H]
\setlength{\tabcolsep}{1.8pt}
\renewcommand{\arraystretch}{1.1}
\centering
\begin{scriptsize}
\begin{tabular}{|rp{2.3in}rrr|}
\hline
& \textbf{Utterances} & \textbf{C1} & \textbf{C2} & \textbf{C3}\\
\hline
\hline
0 & Thank you for sticking around wanna show you some real cool stuff, not right there. & $\square$ & $\square$ & \done\\
1 & It's gotta stay and the same letter so they was the letter C and I'm just using my hands my fingers. & $\square$ & \done & $\square$\\
2 & I do have my Surface Pen somewhere it's up there. & $\square$ & $\square$ & $\square$\\
3 & I'm on the Surface Pro. & \done & $\square$ & $\square$\\
4 & \st{7} & $\square$ & $\square$ & $\square$\\
5 & That's the latest song version of the surface. & $\square$ & $\square$ & $\square$\\
6 & It just came out early in October 20th 19th. & $\square$ & $\square$ & $\square$\\
7 & I know that in a few days were gonna be in 2020, which is why I want to have. & $\square$ & $\square$ & $\square$\\
8 & This file ready an complete I'm just listening. & $\square$ & $\square$ & $\square$\\
9 & Bump on Phone so yeah, that's the wrong letter, but let's go. & $\square$ & $\square$ & $\square$\\
10 & \st{It's just he gone.} & $\square$ & $\square$ & $\square$\\
11 & {\cellcolor[gray]{.88}}So I'm doing this little dots and what happens is we on with Adobe fresco is there. & $\square$ & $\square$ & $\square$\\
12 & It doesn't do a good job. & \done & $\square$ & $\square$\\
13 & {\cellcolor[gray]{.88}}The app and it's working it's a work in progress OK is constantly being updated with the features and bugs. & $\square$ & \done & \done\\
14 & {\cellcolor[gray]{.88}}But I thought we fresco does not have good, um Palm rejection and so when you using the pen and drawing whether you're on the iPad or surface. & $\square$ & \done & \done\\
15 & You end up with all these little artifacts that when your Palm lays on the screen. & $\square$ & $\square$ & $\square$\\
16 & And so as I was drawing diesel letters. & $\square$ & $\square$ & $\square$\\
17 & I got to the point where. & \done & $\square$ & $\square$\\
18 & I just cannot avoid these artifacts in Astana kept them. & $\square$ & $\square$ & $\square$\\
19 & {\cellcolor[gray]{.88}}I kept them in the drawing and now they are part of the look of each of these letters, which is crazy. & $\square$ & \done & $\square$\\
20 & And then there's some of these letters that don't have detail like this one. & $\square$ & $\square$ & $\square$\\
21 & So now I am going to use my pen there. & $\square$ & $\square$ & $\square$\\
22 & We go come on come here pin so I got a Surface Pen here. & $\square$ & $\square$ & $\square$\\
23 & \st{Anthony my glasses.} & $\square$ & $\square$ & $\square$\\
24 & But the lightless terrible OK here we go let me. & $\square$ & $\square$ & $\square$\\
25 & I need to do some of these lines here is geared is closeout all. & $\square$ & $\square$ & $\square$\\
26 & \st{Define.} & $\square$ & $\square$ & $\square$\\
27 & If this doesn't have a pain. & \done & $\square$ & $\square$\\
28 & Come on, it does have a pen. & $\square$ & $\square$ & $\square$\\
29 & \st{Usually.} & $\square$ & $\square$ & $\square$\\
30 & \st{I guess the battery.} & $\square$ & $\square$ & $\square$\\
31 & Just did good thing is I have extra pants, not only do I have all here's one, not only do I have extra pants. & $\square$ & \done & $\square$\\
32 & \st{There we go.} & $\square$ & $\square$ & $\square$\\
33 & I am I have the Surface Pen 4 I have the 3. & $\square$ & $\square$ & $\square$\\
34 & \st{I have.} & $\square$ & $\square$ & $\square$\\
35 & I also have other pins with other brands like the Wacom, the bamboo adanet and this other. & $\square$ & $\square$ & $\square$\\
36 & \st{Pam brand.} & $\square$ & $\square$ & $\square$\\
37 & They sent me uhm this pan to check out. & $\square$ & $\square$ & $\square$\\
38 & \st{They mean the company.} & $\square$ & $\square$ & $\square$\\
39 & At the Renaissance so it's pretty cool. & \done & $\square$ & $\square$\\
40 & OK, here, we go ever it is yeah, some of these. & $\square$ & $\square$ & $\square$\\
41 & {\cellcolor[gray]{.88}}Pan's had details that were missing right now, I'm using the vector pen the vector brushes are in. & $\square$ & $\square$ & \done\\
42 & \st{There we go.} & $\square$ & $\square$ & $\square$\\
43 & I'ma try not to yell but I know I can hear. & $\square$ & $\square$ & $\square$\\
44 & Just a big bounce in my voice and Missy Volume. & $\square$ & $\square$ & $\square$\\
45 & OK, so now we're back we're back 'cause I could hear the sound an it was very loud. & $\square$ & $\square$ & \done\\
46 & I'm sorry I already have a loud voice. & $\square$ & $\square$ & $\square$\\
\hline
\end{tabular}
\end{scriptsize}
\vspace{-0.05in}
\caption{
A transcript snippet from \emph{Creating ABC Childrens Book Art on Adobe Fresco and Adobe Illustrator Part 18}.
We show the most prominent latent codes and their representative utterances (`X').
Human annotated summary utterances are colored gray and ultra-short utterances are crossed out.
}
\label{tab:code3}
\vspace{-0.1in}
\end{table}

\begin{table}[H]
\setlength{\tabcolsep}{5pt}
\renewcommand{\arraystretch}{1.1}
\begin{scriptsize}

\begin{minipage}[b]{0.5\hsize}\centering
\begin{tabular}[t]{|p{2.99in}|}


\hline

\textbf{FluCovRank}\\[1mm]

\textbullet\, learning process proceed / draw this out and post it where people / sit down an just trying to draw a picture / spent like 3 hours / alot like ill draw something like old man i want to draw / consistently draw this character and erase and restart and you cant necessarily / practice will\\[1mm]

\hline
\hline

\textbf{LexRank}\\[1mm]

\textbullet\, The practice will do a lot more for you than drawing a picture.\\[0.5mm]

\textbullet\, You might get one picture that might turn out really cool and you can post it and stuff like that, but if you work on your practice and you get that practice the way that you want it, you can use that practice forever for more than one picture.\\[0.5mm]

\textbullet\, You stop trying to make the picture and get back to practice.\\[0.5mm]

\textbullet\, So if there's things that you need to practice and stuff like that, you stand to gain a lot more from that practice then.\\[0.5mm]

\textbullet\, Sit down an just trying to draw a picture.\\[1mm]

\hline
\hline

\textbf{BART-Large}\\[1mm]

\textbullet\, I do I like to do more practice in anything. Practices gives me this feel that I've never had when I was younger. You know, of course, everybody will tell you. You might get one picture that might turn out really cool and you can post it and stuff like that, but if you work on your practice and you get that practice the way that you want it, you can use that practice forever for more than one picture. So that's generally would always tell myself when I'm drawing, and I'm like. Oh crap, here I go again. I should draw this into a picture and then I end up skipping practice to save whatever I'm doing as a picture so that I can. Just 'cause\\[1mm]

\hline
\hline

\textbf{Quantized Transformer}\\[1mm]

\textbullet\,  Trying to make a picture, right?\\[0.1mm]

\textbullet\,  And then I'm like OK alright I gotta I gotta center myself You know I gotta always gotta check myself every now and again or back that was kind of how it wasn't back then but now I don't really like to draw pictures.\\[0.1mm]

\textbullet\,  I do I like to do more practice in anything.\\[0.1mm]

\textbullet\,  You sit up and you put on like references for the superhero be in movies or something like that.\\[0.1mm]

\textbullet\,  You pull up references on whatever social media or wherever you get your images from.\\[0.1mm]

\textbullet\,  Whereas if you were just trying to practice, you would realize that all of that stuff that you're about to throw away or disregard was actually what you were actually searching for.\\[1mm]

\hline
\hline

\textbf{StreamHover (Ours)}\\[1mm]

\textbullet\, Doesn't matter what styles you draw, so don't overly complicated by like Oh my God, you know now you could draw it.\\[1mm]

\textbullet\, I used to do that a lot when I was younger where I would do something and be like holy crap, didn't think that would come out that way.\\[1mm]

\textbullet\, And then I'm like OK alright I gotta I gotta center myself You know I gotta always gotta check myself every now and again or back that was kind of how it wasn't back then but now I don't really like to draw pictures.\\[1mm]

\textbullet\, They don't see it happening until it's too late where it's like, Oh Man, I could totally practice this right now, or I could Draw Something Super Duper cool, you know, a piece of fan art or something like that.\\[1mm]

\textbullet\, You're going to struggle with for a couple of days or couple minutes and then figure out you can't draw it or you have a hard time drawing it and then just giving up.\\[1mm]

\hline
\end{tabular}
\end{minipage}

\end{scriptsize}
\caption{
Example system summaries from \emph{Call me Derek: Value Study}.
The BART summary is fluent but its content lacks specificity, as is the case for LexRank.
Summary segments selected by FluCovRank are ungrammatical. 
StreamHover identifies on-topic and informative summary utterances.
}
\label{tab:results_output4}
\vspace{-0.1in}
\end{table}

\begin{table}[H]
\setlength{\tabcolsep}{1.8pt}
\renewcommand{\arraystretch}{1.1}
\centering
\begin{scriptsize}
\begin{tabular}{|rp{2.3in}rrr|}
\hline
& \textbf{Utterances} & \textbf{C1} & \textbf{C2} & \textbf{C3}\\
\hline
\hline
0 & {\cellcolor[gray]{.88}}And then once I get my general OK this is what the hairstyle is gonna be right here. & $\square$ & $\square$ & $\square$\\
1 & I definitely want to condom mark the flow. & $\square$ & $\square$ & $\square$\\
2 & Well, I'll do that later. & \done & $\square$ & $\square$\\
3 & And now I just kind of relax and just. & $\square$ & $\square$ & $\square$\\
4 & {\cellcolor[gray]{.88}}Draw hairstyles as long as there as long as his long hair. & $\square$ & $\square$ & $\square$\\
5 & Doesn't matter what styles you draw, so don't overly complicated by like Oh my God, you know now you could draw it. & $\square$ & $\square$ & $\square$\\
6 & \st{Now you can draw.} & $\square$ & $\square$ & $\square$\\
7 & \st{Faces and stuff.} & $\square$ & $\square$ & $\square$\\
8 & If you want, I try not to draw too much of the face. & $\square$ & $\square$ & $\square$\\
9 & Just 'cause I don't want to. & \done & $\square$ & $\square$\\
10 & I don't want to get caught up in that thing that I always do where when I'm drawing a face or I'm doing practice. & $\square$ & $\square$ & $\square$\\
11 & I always do that thing where it's like holy crap you know this is a really good face. & $\square$ & $\square$ & $\square$\\
12 & I should draw this into a picture and then I end up skipping practice to save whatever I'm doing as a picture so that I can. & $\square$ & \done & $\square$\\
13 & \st{You know?} & $\square$ & $\square$ & $\square$\\
14 & Posted on Instagram or something like that. & $\square$ & $\square$ & $\square$\\
15 & {\cellcolor[gray]{.88}}I used to do that a lot when I was younger where I would do something and be like holy crap, didn't think that would come out that way. & $\square$ & $\square$ & \done\\
16 & I don't want to waste this in practice. & \done & $\square$ & $\square$\\
17-21 & \dots & $\square$ & $\square$ & $\square$\\
22 & You might get one picture that might turn out really cool and you can post it and stuff like that, but if you work on your practice and you get that practice the way that you want it, you can use that practice forever for more than one picture. & $\square$ & \done & $\square$\\
23 & So that's generally would always tell myself when I'm drawing, and I'm like. & $\square$ & $\square$ & $\square$\\
24 & Oh crap, here I go again. & $\square$ & $\square$ & $\square$\\
25 & Trying to make a picture, right? & \done & $\square$ & $\square$\\
26 & You stop trying to make the picture and get back to practice. & $\square$ & $\square$ & $\square$\\
27 & {\cellcolor[gray]{.88}}And then I'm like OK alright I gotta I gotta center myself You know I gotta always gotta check myself every now and again or back that was kind of how it wasn't back then but now I don't really like to draw pictures. & $\square$ & \done & \done\\
28 & I do I like to do more practice in anything. & $\square$ & $\square$ & $\square$\\
29 & Practices gives me this feel that I've never had when I was younger. & $\square$ & $\square$ & $\square$\\
30 & When I was younger, which is so eager to get people to see my work that I wasn't necessarily thinking about the overall count, their overall. & $\square$ & $\square$ & \done\\
31 & Affect that, we're going to have on. & $\square$ & $\square$ & $\square$\\
32 & \st{My learning process proceed.} & $\square$ & $\square$ & $\square$\\
33 & And it had a big big, big big like effect on it took me longer to learn stuff because when you sit down you feel like you don't want to do anything unless you're actually drawing something to show off, which is something you don't necessarily want to get caught up in, which happens a lot of people they don't. & $\square$ & \done & $\square$\\
35-36 & \dots & $\square$ & $\square$ & $\square$\\
37 & Get way more out of practice because practice is more focused on getting you better at art where doing a picture or whatever is cool, but it's more focused on showing off what you can do or what you already know. & $\square$ & \done & $\square$\\
38 & So if there's things that you need to practice and stuff like that, you stand to gain a lot more from that practice then. & $\square$ & $\square$ & $\square$\\
39 & \st{You know?} & $\square$ & $\square$ & $\square$\\
40 & Sit down an just trying to draw a picture. & $\square$ & $\square$ & $\square$\\
41 & {\cellcolor[gray]{.88}}You're going to struggle with for a couple of days or couple minutes and then figure out you can't draw it or you have a hard time drawing it and then just giving up. & $\square$ & $\square$ & $\square$\\
42 & Which is something that would happen alot like I'll Draw Something like old man, you know I want to draw this. & $\square$ & $\square$ & $\square$\\
43 & Superhero, I've had a problem with drawing the superhero forever, but today is going to be the day I'm going to nail it. & $\square$ & $\square$ & \done\\
\hline
\end{tabular}
\end{scriptsize}
\vspace{-0.05in}
\caption{
A transcript snippet from \emph{Call me Derek: Value Study}.
We show the most prominent latent codes and their representative utterances (`X').
Human annotated summary utterances are colored gray and ultra-short utterances are crossed out.
}
\label{tab:code4}
\vspace{-0.1in}
\end{table}

\begin{table}[H]
\setlength{\tabcolsep}{5pt}
\renewcommand{\arraystretch}{1.1}
\begin{scriptsize}

\begin{minipage}[b]{0.5\hsize}\centering
\begin{tabular}[t]{|p{2.99in}|}


\hline

\textbf{FluCovRank}\\[1mm]

\textbullet\, people start off always drawing characters / favorite characters from sailor moon / mohammed roller show / yeah people are different or personalities / example prefer environment designed the character design and even though i like looking at character art by prefer to make environments / advice with a grain of salt / people specialized the\\[1mm]

\hline
\hline

\textbf{LexRank}\\[1mm]

\textbullet\, Some people like to have things mixed up all the time where they get bored and other people find that one thing they like and never tire of it.\\[0.5mm]

\textbullet\, Or what I meant is that some people start off always drawing characters, for example like they just from.\\[0.5mm]

\textbullet\, An anyway, some artists just like for example, love drawing characters and always draw them and never get into environments, and they go on and become a good working concepts are it could work in visual development as a character artist who knows what.\\[0.5mm]

\textbullet\, Just because someone specializes, it doesn't necessarily mean that they like tried all the other things first.\\[0.5mm]

\textbullet\, Some people sort of specialized the whole time just because that was their preference, you know.\\[1mm]

\hline
\hline

\textbf{BART-Large}\\[1mm]

\textbullet\, I really works good to see you. Yeah, people are different or personalities. What might work for them isn't necessarily right right for you for any number of reasons, like what David said just from person to person, those preferences can be really different. I for example, prefer environment designed the character design, and even though I like looking at character art by prefer to make environments, and that's not. I think that's why you have to take advice with a grain of salt like. See I do think people do though Omar, like maybe maybe you're not relating to it 'cause it's just not what your preferences are but. Definitely, I think it's important to seek out advice from professional artists and from people\\[1mm]

\hline
\hline

\textbf{Quantized Transformer}\\[1mm]

\textbullet\,  It's about what you like doing.\\[0.1mm]

\textbullet\,  I really works good to see you.\\[0.1mm]

\textbullet\,  A specialist most of the time is someone with more experience.\\[0.1mm]

\textbullet\,  I mean I like for example, I started drawing in like when I was a kid and when I was in middle school and high school was drawn.\\[0.1mm]

\textbullet\,  Just because someone specializes, it doesn't necessarily mean that they like tried all the other things first.\\[1mm]

\textbullet\,  This was the first study.\\[0.1mm]

\textbullet\,  We're doing cloud studies for part of the art club were doing clouds in different scenarios, this time different.\\[0.1mm]

\textbullet\,  This was the second one.\\[0.1mm]

\textbullet\,  See I do think people do though Omar, like maybe maybe you're not relating to it 'cause it's just not what your preferences are but.\\[0.1mm]

\textbullet\,  All thanks claires you thank you.\\[0.1mm]

\textbullet\,  So I do hope that it helps.\\[1mm]

\hline
\hline

\textbf{StreamHover (Ours)}\\[1mm]

\textbullet\, Some people like to have things mixed up all the time where they get bored and other people find that one thing they like and never tire of it.\\[1mm]

\textbullet\, An anyway, some artists just like for example, love drawing characters and always draw them and never get into environments, and they go on and become a good working concepts are it could work in visual development as a character artist who knows what.\\[1mm]

\textbullet\, We're doing cloud studies for part of the art club were doing clouds in different scenarios, this time different.\\[1mm]

\textbullet\, See I do think people do though Omar, like maybe maybe you're not relating to it 'cause it's just not what your preferences are but.\\[1mm]

\textbullet\, I for example, prefer environment designed the character design, and even though I like looking at character art by prefer to make environments, and that's not.\\[1mm]

\hline
\end{tabular}
\end{minipage}

\end{scriptsize}
\caption{
Example system summaries for \emph{Digital Painting Studies with Maddy Bellwoar--Clouds}.
The BART summary is fluent but its content lacks specificity, as is the case for LexRank.
The summary segments selected by FluCovRank are ungrammatical. 
StreamHover identifies on-topic and informative summary utterances.
}
\label{tab:results_output5}
\vspace{-0.1in}
\end{table}

\begin{table}[H]
\setlength{\tabcolsep}{1.8pt}
\renewcommand{\arraystretch}{1.1}
\centering
\begin{scriptsize}
\begin{tabular}{|rp{2.3in}rrr|}
\hline
& \textbf{Utterances} & \textbf{C1} & \textbf{C2} & \textbf{C3}\\
\hline
\hline
0 & Pence Maybe I'll wait on that. & $\square$ & $\square$ & $\square$\\
1 & David says, I also think if you were a person who really likes doing all sorts of different things, you wouldn't be happiness. & $\square$ & $\square$ & \done\\
2 & \st{Specialist role.} & $\square$ & $\square$ & $\square$\\
3 & Yeah, people are different or personalities. & $\square$ & $\square$ & $\square$\\
4 & Some people like to have things mixed up all the time where they get bored and other people find that one thing they like and never tire of it. & $\square$ & \done & \done\\
5 & \st{And it really depends.} & $\square$ & $\square$ & $\square$\\
6 & \st{It really depends.} & $\square$ & $\square$ & $\square$\\
7 & I think that's why you have to take advice with a grain of salt like. & $\square$ & $\square$ & $\square$\\
8 & {\cellcolor[gray]{.88}}Definitely, I think it's important to seek out advice from professional artists and from people that have sort of gone down the path that you're trying to go down. & $\square$ & $\square$ & $\square$\\
9 & {\cellcolor[gray]{.88}}That's going to make things easier for you, but you know, take the advice with a grain of salt because. & $\square$ & $\square$ & $\square$\\
10 & What might work for them isn't necessarily right right for you for any number of reasons, like what David said just from person to person, those preferences can be really different. & $\square$ & \done & $\square$\\
11 & Exactly Pablo is at the end of the day. & $\square$ & $\square$ & $\square$\\
12 & It's about what you like doing. & \done & $\square$ & $\square$\\
13 & Alright, I'm gonna grab a little more textured brush again. & $\square$ & $\square$ & $\square$\\
14 & {\cellcolor[gray]{.88}}I'm kind of going back and forth with harder brushes and textured brushes. & $\square$ & $\square$ & $\square$\\
15 & I really works good to see you. & $\square$ & $\square$ & $\square$\\
16 & \st{It's not about that.} & $\square$ & $\square$ & $\square$\\
17 & A specialist most of the time is someone with more experience. & $\square$ & $\square$ & $\square$\\
18 & That's why he's a specialist. & \done & $\square$ & $\square$\\
19 & Oh, I understand your thoughts on that. & $\square$ & $\square$ & $\square$\\
20 & I would agree for the most part, but I guess the part I would disagree on is that. & $\square$ & $\square$ & $\square$\\
21 & {\cellcolor[gray]{.88}}Or what I meant is that some people start off always drawing characters, for example like they just from. & $\square$ & $\square$ & $\square$\\
22 & I mean I like for example, I started drawing in like when I was a kid and when I was in middle school and high school was drawn. & $\square$ & $\square$ & \done\\
23 & All my favorite characters from Sailor Moon and all that kind of stuff. & $\square$ & $\square$ & $\square$\\
24 & {\cellcolor[gray]{.88}}An anyway, some artists just like for example, love drawing characters and always draw them and never get into environments, and they go on and become a good working concepts are it could work in visual development as a character artist who knows what. & $\square$ & \done & \done\\
25 & Just because someone specializes, it doesn't necessarily mean that they like tried all the other things first. & $\square$ & $\square$ & $\square$\\
26 & \st{Maybe they did.} & $\square$ & $\square$ & $\square$\\
27 & It's possible they did, but I don't think that it has to be that you have to do everything before you specialize. & $\square$ & \done & $\square$\\
28 & Some people sort of specialized the whole time just because that was their preference, you know. & $\square$ & $\square$ & $\square$\\
29 & \st{From the beginning.} & $\square$ & $\square$ & $\square$\\
30 & If anybody is just coming in. & $\square$ & $\square$ & $\square$\\
31 & Mohammed roller, I can show you what we did so far. & $\square$ & $\square$ & $\square$\\
32 & This was the first study. & \done & $\square$ & $\square$\\
33 & {\cellcolor[gray]{.88}}We're doing cloud studies for part of the art club were doing clouds in different scenarios, this time different. & $\square$ & $\square$ & $\square$\\
34 & Like lightings, different times of day. & $\square$ & $\square$ & $\square$\\
35 & This was the second one. & \done & $\square$ & $\square$\\
36 & \st{And.} & $\square$ & $\square$ & $\square$\\
37 & This is what we're working on right now. & \done & $\square$ & $\square$\\
38 & See I do think people do though Omar, like maybe maybe you're not relating to it 'cause it's just not what your preferences are but. & $\square$ & $\square$ & $\square$\\
39 & {\cellcolor[gray]{.88}}I for example, prefer environment designed the character design, and even though I like looking at character art by prefer to make environments, and that's not. & $\square$ & \done & \done\\
40 & You know, that's just my preference. & $\square$ & $\square$ & $\square$\\
41 & All thanks claires you thank you. & $\square$ & $\square$ & $\square$\\
\hline
\end{tabular}
\end{scriptsize}
\vspace{-0.05in}
\caption{
A snippet from \emph{Digital Painting Studies with Maddy Bellwoar--Clouds}.
We show the most prominent latent codes and their representative utterances (`X').
Human annotated summary utterances are colored gray and ultra-short utterances are crossed out.
}
\label{tab:code5}
\vspace{-0.1in}
\end{table}

\end{document}